\RequirePackage{fix-cm}
\documentclass[twocolumn,natbib]{svjour3}           %
\smartqed  %
\usepackage{graphicx}
\journalname{International Journal of Computer Vision}

\usepackage{url}
\usepackage{booktabs}       %
\usepackage{amsfonts}       %
\usepackage{algorithm} 
\usepackage{algpseudocode}  %
\usepackage{nicefrac}       %
\usepackage{microtype}      %
\usepackage{graphicx}		%
\usepackage{pgfplots}
\usepackage{pgf}            
\usepackage{tikz}                   
\usetikzlibrary{arrows,shapes,snakes,
               automata,backgrounds,
               petri,topaths}
\usepackage{subcaption}     %
\usepackage{wrapfig}        %
\usepackage{makecell}       %
\usepackage{multirow}       %
\usepackage{mdwtab,booktabs}%
\usepackage{float}          %
\usepackage{amsmath}
\usepackage{bm}             %
\usepackage{dsfont}         %
\usepackage[export]{adjustbox} %
\usepackage{soul}           %
\usepackage{rotating}
\usepackage[colorlinks=true, allcolors=blue]{hyperref}
\usepackage[utf8]{inputenc}
\newcommand\EE{\mathbb{E}}
\newcommand{\FGen}{T}
\newlength{\twosubht}
\newsavebox{\twosubbox}
\newcommand{\R}{\mathbb{R}}
\newcommand{\PJ}[2]{\mathbb{P}_{#1#2}}  %
\newcommand{\PI}[2]{\mathbb{P}_{#1}\otimes\mathbb{P}_{#2}}  %

\def\makeheadbox{{%
\hbox to0pt{\vbox{\baselineskip=10dd\hrule\hbox
to\hsize{\vrule\kern3pt\vbox{\kern3pt
\hbox{\bfseries International Journal of Computer Vision}
\hbox{This is a pre-peer-review pre-print of an article published in the above journal.}
\hbox{The final authenticated version is available online at: \href{[http://link.springer.com/article/10.1007/s11263-020-01322-1]}{DOI 10.1007/s11263-020-01322-1}.}
\kern3pt}\hfil\kern3pt\vrule}\hrule}%
\hss}}}

\begin{document}

\title{Pix2Shape -- Towards Unsupervised Learning of 3D Scenes from Images using a View-based Representation%
}
\subtitle{}

\titlerunning{Pix2Shape}        %

\author{Sai Rajeswar$^{1,2}$ \and
        Fahim Mannan$^{3}$ \and
        Florian Golemo$^{1}$ \and
        Jérôme Parent-Lévesque$^{1}$ \and
        David Vazquez$^{2}$ \and
        Derek Nowrouzezahrai$^{4}$ \and
        Aaron Courville$^{1}$
}

\authorrunning{Sai Rajeswar \etal} %

\institute{Sai Rajeswar \at
              \email{rajsai24@gmail.com}           %
        \and
              $^1$Universit\'e de Montreal, Montreal, Canada
        \and
              $^2$Element AI, Montreal, Canada
        \and
              $^3$Algolux, Montreal, Canada
        \and
              $^4$McGill University, Montreal, Canada
}

\date{Originally submitted: 15th May 2019}

\maketitle

\begin{abstract}
We infer and generate three-dimensional (3D) scene information from a single input image and without supervision. This problem is under-explored, with most prior work relying on supervision from, e.g., 3D ground-truth, multiple images of a scene, image silhouettes or key-points. We propose \textbf{Pix2Shape}, an approach to solve this problem with four component: (i) an encoder that infers the latent 3D representation from an image, (ii) a decoder that generates an explicit 2.5D surfel-based reconstruction of a scene -- from the latent code -- (iii) a differentiable renderer that synthesizes a 2D image from the surfel representation, and (iv) a critic network trained to discriminate between images generated by the decoder-renderer and those from a training distribution. Pix2Shape can generate complex 3D scenes that scale with the view-dependent on-screen resolution, unlike representations that capture world-space resolution, i.e., voxels or meshes. We show that Pix2Shape learns a consistent scene representation in its encoded latent space, and that the decoder can then be applied to this latent representation in order to synthesize the scene from a novel viewpoint.  We evaluate Pix2Shape with experiments on the ShapeNet dataset as well as on a novel benchmark we developed -- called 3D-IQTT -- to evaluate models based on their ability to enable 3d spatial reasoning. Qualitative and quantitative evaluation demonstrate Pix2Shape's ability to solve scene reconstruction, generation and understanding tasks.

\end{abstract}
\section{Introduction}
\label{sec:introv2}

Humans sense, plan and act in a 3D world despite only directly observing 2D projections of their 3D environment. Automatic 3D understanding seeks to recover a realistic underlying 3D structure of a scene using only 2D image projection(s). This long-standing challenge in computer vision has recently admitted learning-based solutions. Many such approaches leverage 3D supervision, such as from images annotated with ground truth 3D shape information~\citep{Girdhar2016, Wu2015, Wu2016, choy2016}. Recent approaches rely on using other forms of 3D supervision, such as multiple views of the same object~\citep{Xinchen2016, TulsianiZEM17, depth2019}, 2.5D supervision~\citep{3DINN, marrnet}, key-point~\citep{Kar14, C3DPOC3} and silhouette annotations~\citep{silnet, henderson18,chen18}. Our work treats the problem of \textit{unsupervised single image 3D scene understanding}. This form of the problem is challenging, as we aim to infer an encoding of 3D structure from only a \textit{single image}, and this too without any form of 3D ground truth supervision during training. %
We do not rely on any 3D scene supervision, however we employ camera pose, scene reflectance profiles and outgoing/observed radiance as weak supervision signals.

\begin{figure*}[t]
\sbox\twosubbox{%
  \resizebox{\dimexpr.95\textwidth-1em}{!}{%
    \includegraphics[height=1.7cm]{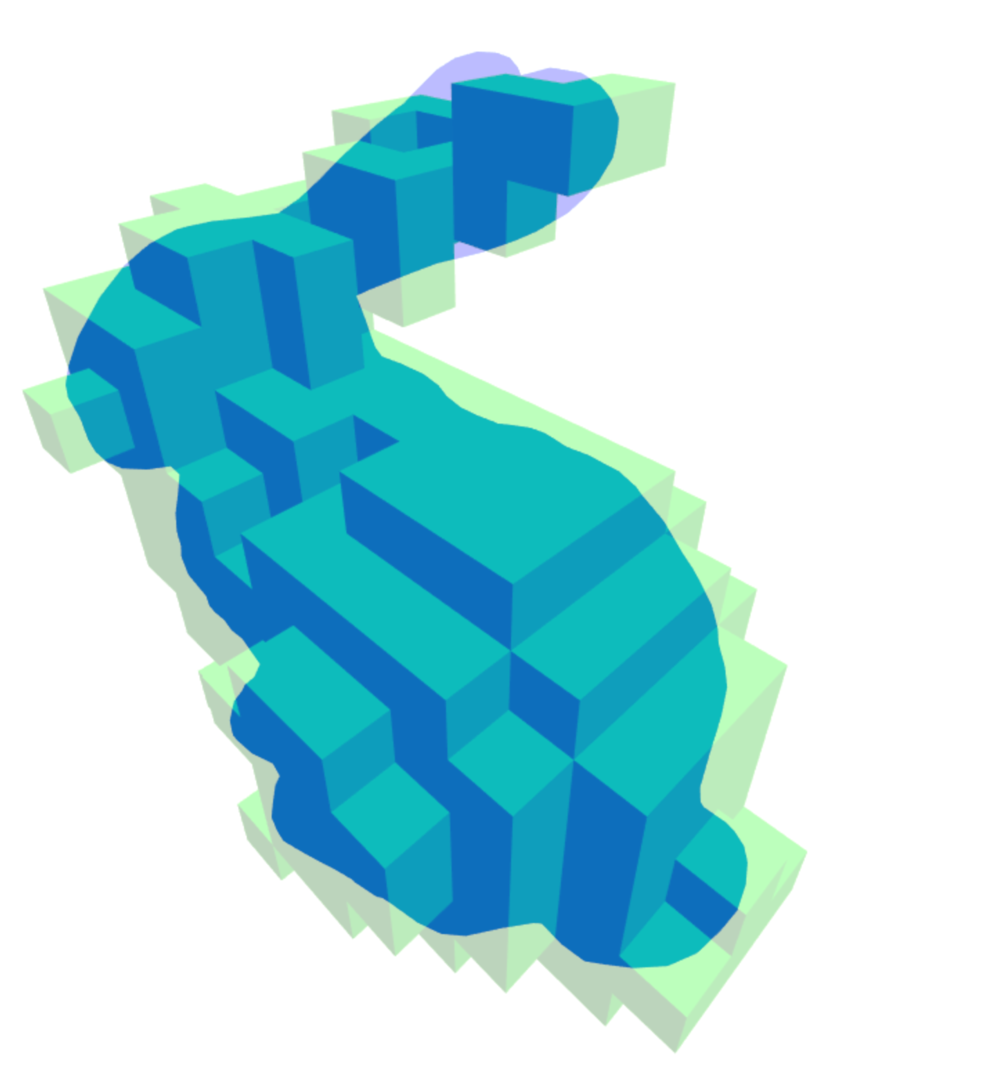}%
    \includegraphics[height=1.7cm]{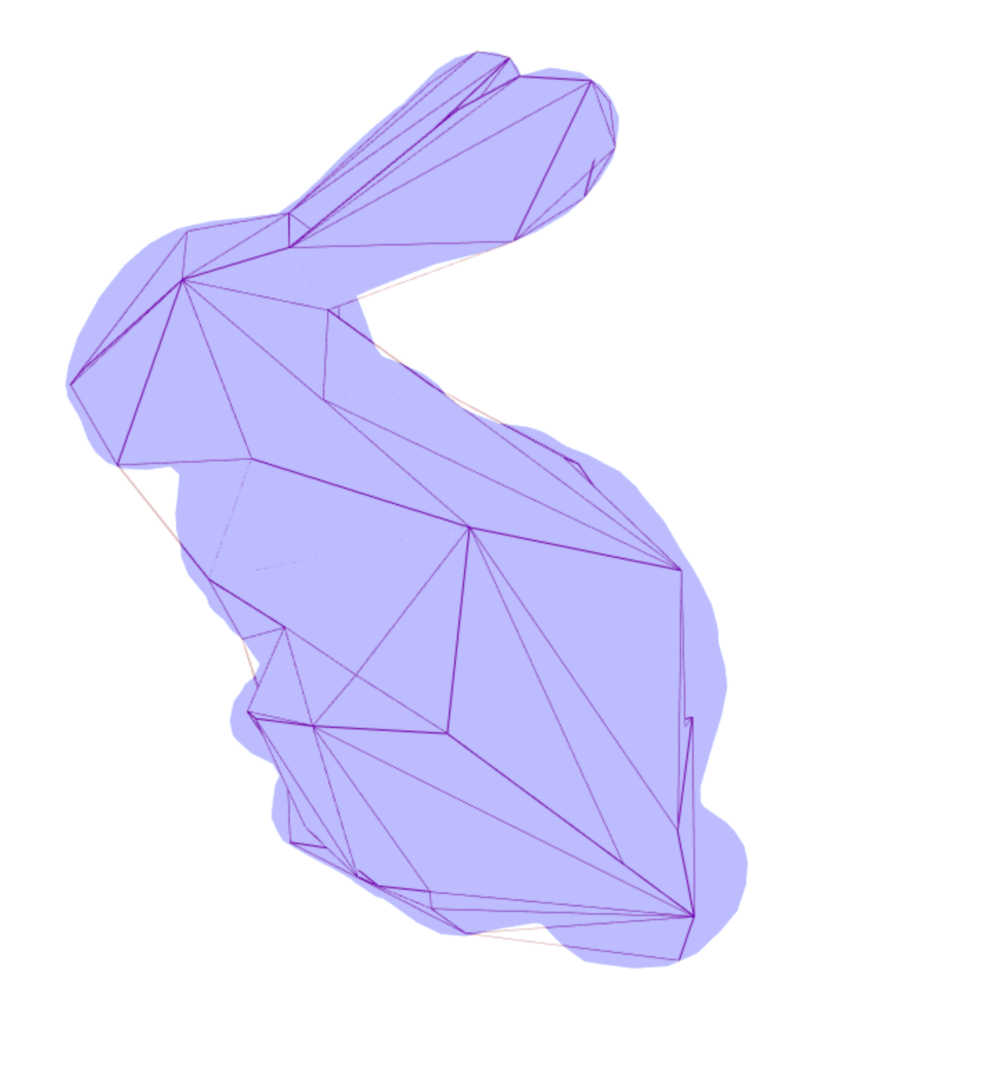}%
    \includegraphics[height=1.7cm]{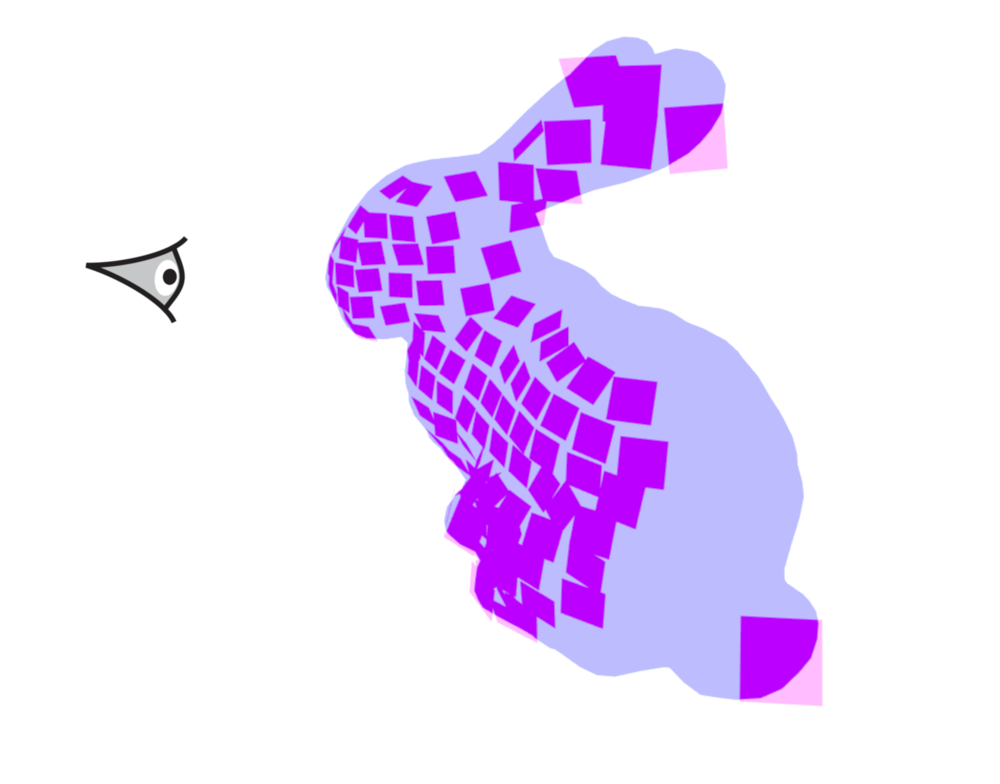}%
    \includegraphics[height=1.7cm]{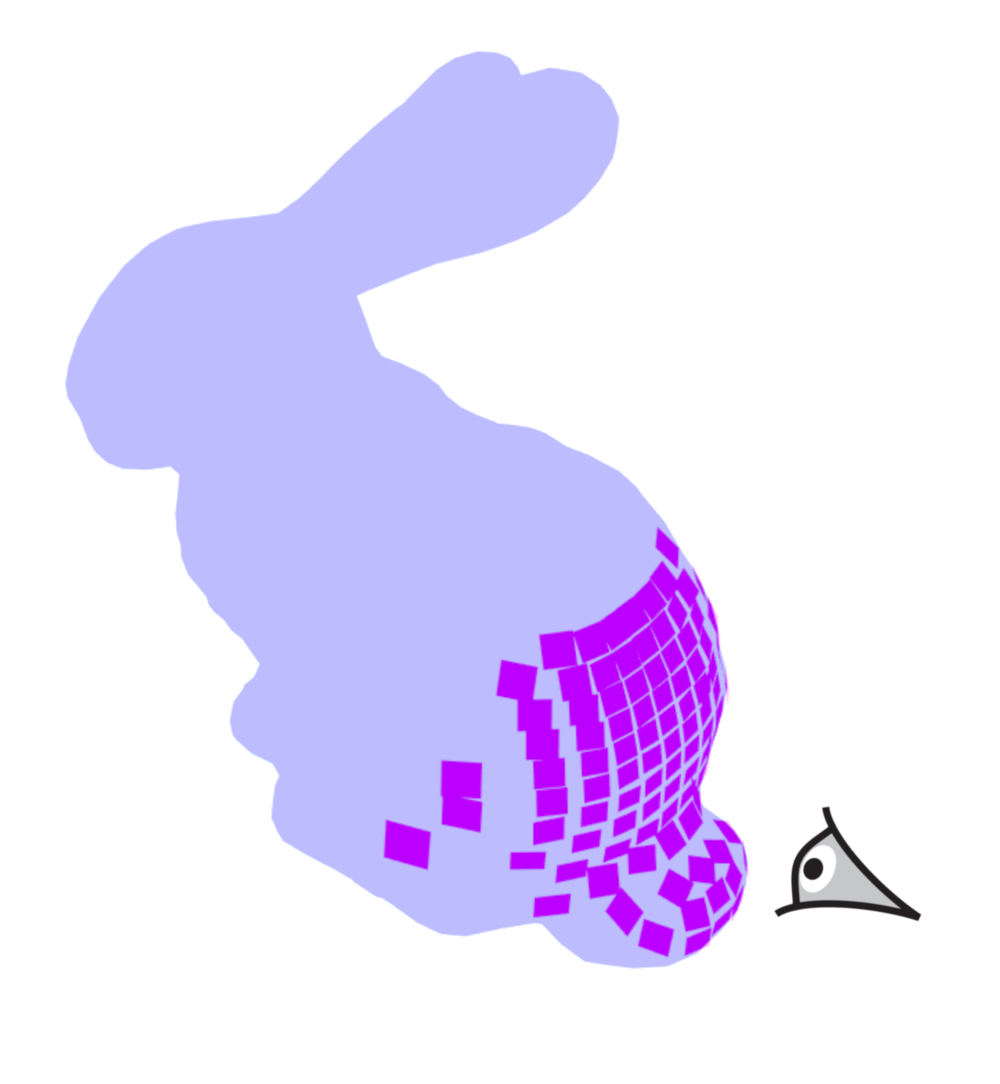}%
  }%
}
\setlength{\twosubht}{\ht\twosubbox}

\centering
\subcaptionbox{Voxel\label{fig:repra}}{%
  \includegraphics[height=\twosubht]{./fig1a.png}%
}\quad
\subcaptionbox{Mesh\label{fig:reprb}}{%
  \includegraphics[height=\twosubht]{./fig1b.png}%
}\quad
\subcaptionbox{Surfels (distant camera)\label{fig:reprc}}{%
  \includegraphics[height=\twosubht]{./fig1c.png}%
}\quad
\subcaptionbox{Surfels (up close)\label{fig:reprd}}{%
  \includegraphics[height=\twosubht]{./fig1d.png}%
}
\caption{\textbf{Comparison of 3D representations.} Voxels and meshes (\ref{fig:repra} and \ref{fig:reprb}) are viewpoint-independent representations. These representations require storage space proportional to the required level of detail. Our implicit representation captures the full scene in a fixed-length latent vector, which, given a viewpoint, can be decoded into an explicit viewpoint-dependent ``surfels" representation with arbitrary level of detail (\ref{fig:reprc} and \ref{fig:reprd}).}
\label{fig:implicitVSexplicit}
\end{figure*}

While the benefits of employing supervision can certainly be argued for in this context -- i.e., with the growing number of datasets with labelled 3D ground truth for objects~\citep{shapenet2015} and cityscapes~\citep{nuscenes2019} -- one benefit of approaching the problem from an unsupervised perspective is that we are not limited to the types of 3D objects represented in these datasets. Indeed, however vast, existing datasets fall far from capturing all possible artificial and natural 3D scenes and objects. Moreover, datasets with depth annotations often contain incomplete or noisy depth maps due to limitations in depth capture hardware.

Unsupervised single image 3D understanding is a relatively under-explored area, with only a few works treating this setting~\citet{Rezende2016, Xinchen2016}. These methods rely on deformable 3D mesh or voxel representations of the world, and have only been applied to simple 3D primitives (e.g., cubes, spheres) or single objects over a clean background. 

One approach to this problem is to leverage prior knowledge on how 2D images are formed from the 3D world, including the effects of shading and occlusion. Building machine learning architectures with an explicit knowledge of this \textit{forward rendering} model could help better disambiguate the 3D structure of geometry from 2D observations. In this spirit, we propose the \textbf{Pix2Shape} architecture for unsupervised single image 3D understanding: a model that learns abstract latent encodings of the geometry of an entire scene geometry, and all from a single image. These implicit learnt scene representations can be decoded -- when combined with a targe viewing/camera position -- into a view-dependent realization of 2.5D surfaces (depth map and surface normals) visible only from that view. We can then readily re-render these explicit view-dependent surface elements (surfels) at their corresponding 2D image projections in order to synthesize an unseen view of the scene.

Our model builds atop Adversarially Learned Inference (ALI)~\citep{dumoulin2016adversarially}, an extension of Generative Adversarial Networks (GANs)~\citep{Goodfellow2014} that infers a latent code from an image using an encoder network. In Pix2Shape, the encoder network learns a latent representation that embeds the 3D information of \textit{an entire scene} from an image. We map the latent representation to view-dependent depth and normal maps using a decoder before projecting these maps onto image space using a differentiable renderer. We evaluate the resulting image using an adversarial critic. Our model remains unsupervised as it does not require ground truth depth maps nor any other kind of 3D supervision, as in previous works (e.g., observing the same object from multiple views, key-point registration or image silhouettes). Note that, at any given instant, our model outputs the depth and surface normals conditioned on a specific camera view; we never produce/synthesize the entirety of the 3D world structure. That being said, the latent space we learn embeds the 3D geometry of the entire underlying scene, which allows our decoder and renderer to smoothly extrapolate and synthesize scene geometry from unseen camera views during inference. We refer to this indirect process of embedding 3D information in the latent code as ``\textit{implicit}'' inference.

An ambitious long-term goal is to infer the 3D structure of photographs of the real-world, and our work takes a first step in this direction: We rely on physically based rendering in-order to build a model of the world. However, in order to make the training tractable we experiment exclusively with synthetically constructed scenes, adopting several simplifying assumptions. Of note, we assume that the world is composed of piece-wise smooth 3D elements and that, for each input image, the illumination, view and object materials are known. Since each pixel in an image is a function of geometry, illumination, view and texture, our focus in this work is to learn the underlying geometry of a scene keeping the other parameters fixed.

\begin{figure*}[t]
    \centering
    \includegraphics[width=0.49\textwidth]{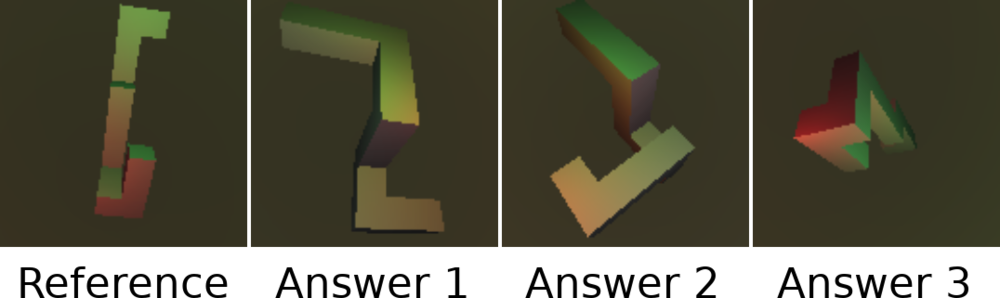}\hfill
    \includegraphics[width=0.49\textwidth]{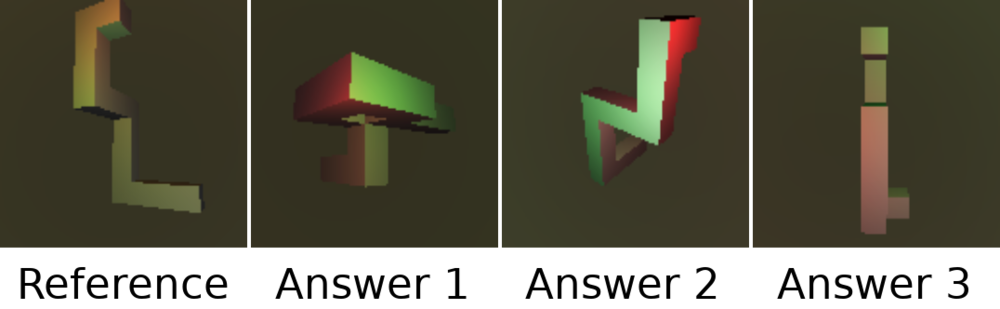}
    \caption{\textbf{Sample questions from 3D-IQTT.} For this ``mental rotation" task, we present a reference image and three possible answers. The test is a classification task where the goal is to find the rotated view of the model from the reference image. To solve this task, the 3D shape of the reference must be inferred from the 2D image and compared to the inferred 3D shapes of the answers (see footnote for correct answers).}
    \label{fig:iqtest-sample}
\end{figure*}

We evaluate our model's ability to recover accurate and consistent depth from a single image, for both seen and unseen viewpoints, using Hausdorff and Chamfer distance metrics between generated and ground truth depth maps. In addition to reconstruction, we can sample novel scenes (at novel views) using the generative nature of our adversarial network. Finally, we propose a new 3D understanding benchmark -- \textit{3D IQ Test Task} ({3D-IQTT}) -- to evaluate models' understanding of the underlying 3D structure of an object: the test consists of matching a rotated view of a reference object (Figure~\ref{fig:iqtest-sample}). To perform this task, we develop a novel 3D-IQ dataset to train and test against. In this setting, we can additionally estimate camera pose in our learnt latent 3D world embedding. Our contributions are as follows:
\begin{enumerate}
    \item an approach for unsupervised single image 3D understanding that builds a latent embedding of an entire 3D scene,
    \item a decoding scheme that leverages view-dependent, explicit surfel representations to sample scene information more efficiently than (world-space) voxels and meshes,
    \item a differentiable 3D renderer that we leverage, and that can be included as a layer in any learning-based neural network architecture, and
    \item 3D-IQTT, a new 3D understanding benchmark.
\end{enumerate}

\section{Related Work}
\label{sec:related_work}
\subsection{Single view 3D Reconstruction and Generation }
3D generation and reconstruction has been studied extensively in the computer vision and graphics communities~\citep{Saxena:make3d, Chaudhuri2011, Kalogerakis2012, shapenet2015, Rezende2016, Soltani2017, Kulkarni2015, tulsiani16,framenet,jiang19}. Most methods in the literature focus on recovering the 3D structure from 2D images by using explicit 3D supervision. \citet{choy2016, Girdhar2016, Wu2016, Wu2015, VON} reconstruct and/or generate 3D voxels from a latent representation by directly comparing with available 3D shapes. \citet{marrnet, Zhang2018} use 2.5D supervision during training, i.e., depth maps. More recent methods tend to use weaker forms of supervision for single image reconstruction.  \citet{3DINN, kato18, henderson18, chen18} use image based annotations like silhouettes, 2D keypoints or object masks. \citet{cmrKanazawa18} learn both texture and shape from 2D images leveraging multiple learning signals such as keypoints and mean shape. 

\citet{Rezende2016, Xinchen2016,GadelhaMW16,NovotnyLV17a} learn 3D shapes by using multiple views and approximately differentiable rendering mechanisms. However, one of \citet{Rezende2016}'s experiments show reconstruction 3D objects trained using a single view. As far as we know, theirs is the only fully unsupervised method for explicit 3D reconstruction from a single image. Their method is limited to reconstructing relatively simple 3D primitives floating in space due to the strong priors required for the model to work. Concurrent to our work, HoloGAN~\citep{hologan} can synthesize 2D images of more realistic scenes (e.g., cars, bedrooms) under camera view rotation. However their model can not recover the geometry from its implicit representation. Compared to \citet{Rezende2016}, our model can learn to represent more complex synthetic indoor scenes composed of multiple ShapeNet\citep{shapenet2015} objects and, while we do not address real image inputs (i.e., as  HoloGAN), we can infer explicit geometry for visible surfaces from each given view. As such, our model can also be applied to 3D reconstruction (like \citet{Rezende2016} but only for visible parts of the scene) and novel viewpoint image generation (like \citet{hologan}).

\subsection{Differentiable Rendering}
In order to facilitate deep neural network based models to infer 3D structures from their 2D projections (images), it is required to compute and propagate the derivatives of image pixels with respect to 3D geometry and other properties. Gradient estimation through rendering process is a challenging task. In both rasterization and ray-tracing techniques the visibility mapping step is often non-differentiable. \citet{opendr} is one of the well known methods for differential rendering, but has limited applicability due to high computational and memory costs. \citet{kato17, Rezende2016} approximate the gradients of the rendering process and are often limited to a rasterization based rendering scheme. OpenDR~\citep{opendr}, as used by \citet{henderson18}, applies first order Taylor approximation to compute gradients. \citet{Liu2019} computes the gradients analytically by softly assigning contribution of each triangle face to a pixel in mesh-based representations. \citet{chen18} improved this soft assignment and allow the use of textures by interpolating local mesh properties for foreground pixels. \citet{eldar2018} proposed a differentiable re-projection mechanism for point clouds to infer 3D shapes. However learning methods built on these approaches so-far require either more than one view per object or 2D silhouette as supervision and can only reconstruct single objects. In our work we circumvent the non-differentiablity challenge as follows: (1) Our network is trained to output only ``visible" surface elements (surfels) of the scene conditioned on the view, i.e. a 2.5D representation and (2) We maintain one-to-one correspondence between the output surfels and the pixels. In other words our model outputs exactly one surfel in object space per pixel in the output image, and the final image is then formed by a differentiable shading operation. This makes our model differentiable, easily adaptable across image resolutions and allows end-to-end training.

\renewcommand{\thefootnote}{}
\footnotetext{\raggedleft\rotatebox{180}{$^1$ three\qquad $^2$ two\hfill}}

\section{Method}
\label{sec:method}

\begin{figure*}[t]
  \centering
  \includegraphics[width=.99\textwidth]{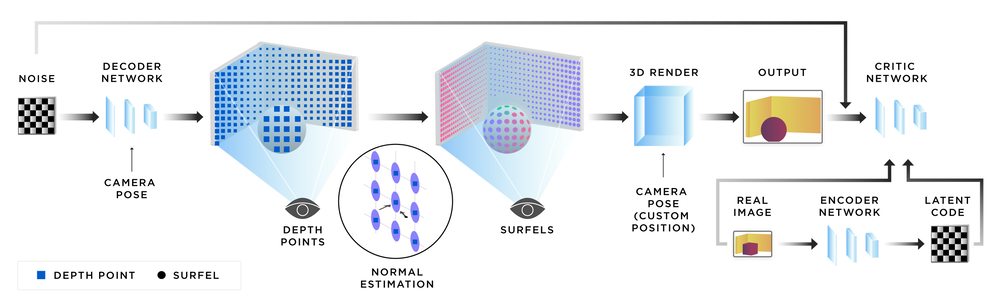}
  \caption{\textbf{Model.} Pix2Shape generates realistic 3D views of scenes by training on 2D single images only. Its decoder generates the surfel depth map $p_z$ from a noise vector $\bm{z}$ conditioned on the camera pose. The surfel normals are estimated from the predicted depth. The surfels are rendered into a 2D image and, together with image samples from the target distribution, are fed to the critic, which generates a gradient for both encoder and decoder paths.}
  \label{fig:model}
\end{figure*}

Our method follows the ALI architecture~\citep{dumoulin2016adversarially}, where we have an encoder branch that learns to convert images into latent representations, a decoder branch that learns to generate images from randomly sampled latent representations, and a critic that tries to predict if pairs of latent code and image are real or fake. The critic and encoder pathways are implemented as convolutional neural networks but the decoder pathway contains an additional differentiable renderer, usable like a layer of a neural network, that converts the 2.5D surfel representation into a 2D image by computing shading at each surfel. Additionally, the decoder is conditioned on a camera pose. See Figure~\ref{fig:model} for an overview. In the following section, we drill down on the individual components of this architecture.

\subsection{3D Representation and Surfels}
\label{subsec:method_surfels}
Representing 3D structure as voxels or meshes presents different challenges for generative models~\citep{kobbelt2004survey}. Representing entire objects using voxels scales poorly given its $O(n^3)$ complexity. Additionally, the vast majority of the generated voxels are not relevant to most viewpoints, such as the voxels that are entirely inside objects. A common workaround is to use a surface representation such as meshes. However, these too come with their own drawbacks, such as their graph-like structure. This makes mesh representation difficult to generate using neural networks. Current mesh based methods mainly rely on deforming a pre-existing mesh and are thus limiting the object topology to have the same genus as the template mesh.

Our approach represents the 3D scene implicitly in a high-dimensional latent variable. In our framework, this latent variable (i.e., a vector) is decoded using a decoder network conditioned on the camera pose into a viewpoint-dependent representation of surface elements (i.e., surfels~\citep{Pfister}, square-shaped planes that are scaled based on depth to roughly fit the size of a pixel) that constitute the visible part of the scene. This representation is very compact: given a renderer's point of view, we can represent only the part of the 3D surface needed by the renderer. As the camera moves closer to a part of the scene, surfels become more compact and thereby increase the amount of visible detail. For descriptive purpose we discuss surfels as squares, but in general they can have any shape. Figure~\ref{fig:implicitVSexplicit} compares surfels with different representations. Surfels differ from other explicit representations in that they are view-dependent, i.e., this representation changes for different camera poses (but the implicit latent vector representation does not).

Formally, surfels are represented as a tuple $(P, N, \rho)$, where $P = (p_x, p_y, p_z)$ is its 3D position, $N = (n_x, n_y, n_z)$ is the surface normal vector, and $\rho = (k_r, k_g, k_b)$ is the albedo of the surface material. Note that $\rho$ represents the material properties at the point $P$ and could take a different size for a different shading model. Since we are only interested in modelling structural properties of the scenes, i.e. geometry and depth, we assume that objects in the scene have uniform material properties and thus keep $\rho$ fixed. We also estimate the normals from depth by assuming locally planar surfaces. We represent the surfels in the camera coordinate system and generate one surfel for each pixel in the output image. This makes our representation very compact. Thus, the only necessary parameter for the decoder network to generate is $p_z$, i.e. a depth map.

\subsection{Differentiable 3D Renderer}
\label{subsec:method_renderer}
Since our architecture is GAN-like and uses 2D images as input to the critic network, we need to project the generated 3D representations down to 2D space using a renderer. In our setting, each stage of the rendering pipeline must be differentiable to allow us to take advantage of gradient-based optimization and backpropagate the critic's error signal to the surfel representation. Our proposed rendering process is differentiable because: (1) each output pixel depends exactly on one surfel, and (2) we employ a differentiable shading operation to compute the color of each pixel. 
Our PyTorch implementation of the differentiable renderer can render a $128 \times 128$ surfel-based scene in under 1.4 ms on a mobile NVIDIA GTX 1060 GPU. Further details about the rendering implementation can be found in appendix \ref{app:rendering}.

\subsection{Model}
\label{sec:model_generative}
The adversarial training paradigm allows the generator network to capture the underlying target distribution by competing with an adversarial critic network. Pix2Shape employs bi-directional adversarial training~\citep{dumoulin2016adversarially, donahue2016adversarial} to model the distribution of surfels from 2D images.

\subsubsection{Bi-Directional Adversarial Training}
ALI~\citep{dumoulin2016adversarially} or Bi-GAN~\citep{donahue2016adversarial} extend the GAN~\citep{Goodfellow2014} framework by including the learning of an inference mechanism. Specifically, in addition to the decoder network $G_x$, ALI provides an encoder $G_z$ which maps data points $\bm{x}$ to latent representations $\bm{z}$. In these bi-directional models, the critic, $D$, discriminates in both the data space ($\bm{x}$ versus $G_x(\bm{z})$), and latent space ($\bm{z}$ versus $G_z(\bm{x})$) jointly, maximizing the adversarial value function over two joint distributions. The final min-max objective can be written as:
\begin{equation}
\begin{split}
\begin{aligned}
\label{eq:ali_shading}
 \min_{G}\max_{D} \mathcal{L}_{ALI}(G,D) \!=~ &\mathds{E}_{q(\bm{x})}[\log (D(\bm{x},G_z(\bm{x})))]\\ &+ \mathds{E}_{p(\bm{z})}[\log (1-D(G_x(\bm{z}),\bm{z}))],\notag
\end{aligned}
\end{split}
\end{equation}
where $q(\bm{x})$ and $p(\bm{z})$ denote encoder and decoder marginal distributions.

\subsubsection{Modelling Depth and Constrained Normal Estimation}
The encoder network captures the distribution over the latent space of the scene given an image data point $\bm{x}$. The decoder network maps a fixed scene latent distribution $p(\bm{z}_{scene})$ (a standard normal distribution in our case) to the 2.5D surfel representation from a given viewpoint $\bm{z}_{view}$. The surfel representation is rendered into a 2D image using our differentiable renderer. The resulting image is given as input to the critic to distinguish it from the ground truth image data. To emphasize on the notation, note that the output of the encoder is $\bm{z}_{scene}$ and the input to decoder is $(\bm{z}_{scene},\bm{z}_{view})$

A straightforward way to design the decoder network is to learn a conditional distribution to produce the surfels' depth ($p_z$) and normal ($N$).
However, this could lead to inconsistencies between the local shape and the surface normal. For instance, the decoder can fake an RGB image of a 3D shape simply by changing the normals while keeping the depth fixed. To avoid this issue, we exploit the fact that real-world surfaces are locally planar, and that surfaces visible to the camera have normals constrained to be in the half-space of visible normal directions from the camera's view point. Considering the camera to be looking along the $-z$ axis direction, the estimated normal has the constraint $n_z > 0$. Therefore, the local surface normal is estimated by solving the following problem for every surfel:
\begin{equation}
\label{eq:normal_estimation}
 N^T \nabla P = 0 \mbox{~~subject to~~} \|N\| = 1 \mbox{~~and~~} n_z > 0,
\end{equation} 
where the spatial gradient $\nabla P$ is computed for each of the 8 neighbour points, and $P$ is the position of the surfels in the camera coordinate system obtained by back-projecting the generated depth along rays. 
This approach enforces consistency between the predicted depth field and the computed normals and provides a gradient signal to the depth from the shading process. If the depth is incorrect, the normal-estimator outputs an incorrect set of normals, resulting in an inconsistent RGB image with the data distribution, which in turn would get penalized by the critic. The decoder network is thus incentivized to produce realistic depths.

\subsubsection{Unsupervised Training}
The Wasserstein-GAN~\citep{Arjovsky2017} formulation provides stable training dynamics using the first Wasserstein distance between the distributions. We adopt the gradient penalty setup as proposed in \citet{Gulrajani2017} for more robust training. However, we modify the formulation to take into account the bidirectional training. 

The architectures of our networks, and training hyper-parameters are explained in detail in the supplementary material section~\ref{app:architecture}. Briefly, we used Conditional Normalization~\citep{dumoulin2016adversarially, perez2017film} for conditioning the viewpoint (or camera pose) in the encoder, decoder and the discriminator networks. The viewpoint is a three dimensional vector representing positional coordinates of the camera. In our training, the affine parameters of the batch-normalization layers~\citep{ioffe2015batch} are replaced by learned representations based on the viewpoint. The final objective includes a bi-directional reconstruction loss:
\begin{equation}
\centering
\begin{split}
\begin{aligned}
\label{eq:rec_loss}
 \mathcal{L}_{recon} =~ &\mathds{E}_{q(\bm{x})}[||\bm{x}- \mbox{\textsc{rend}}(G_{x}(G_{z}(\bm{x})))||_2] + \\ &\mathds{E}_{p(\bm{z})}[||\bm{z}- G_{z}(\mbox{\textsc{rend}}(G_{x}(\bm{z})))||_2],
\end{aligned}
\end{split}
\end{equation}
where the \mbox{\textsc{rend}}$(\cdot)$ function synthesizes images through view-dependent decoding and projection and $\bm{z}$ is $(\bm{z}_{scene},\bm{z}_{view})$. This objective enforces the reconstructions from the model to stay close to the corresponding inputs. This reconstruction loss is used for the encoder and decoder networks as it has been empirically shown to improve reconstructions in ALI-type models~\citep{li2017alice}.
\begin{algorithm}[t]
  \caption{Semisupervised classification}
  \begin{algorithmic}[1]
   
    \While {$iter < max\_iter$}
      \State $D \gets$ MiniBatch$()$
      \State $z_x \sim Enc(\bm{x}) ; \forall x \in \{\bm{x}_{ref}, \bm{x}_{d_1}, \bm{x}_{d_2}, \bm{x}_{ans}\} \in D $
      \State $L \gets$ $\mathcal{L}_{ALI} + \mathcal{L}_{recon} + I_{\Theta}(z_{scene}, z_{view})$ 
      \If {supervised-training-interval$(iter)$}
          \State $L \gets L + \mathcal{L}_{\theta}$ %
      \EndIf
      \State optimize networks with $L$
    \EndWhile
\end{algorithmic}
\label{alg:semi-training}
\end{algorithm}

\subsubsection{Semi-supervised Training for Classification}
\label{3diqtt_method}
Our model can be also trained in a semi-supervised setting (see Algorithm~\ref{alg:semi-training}) for solving image classification tasks that require 3D understanding such as the 3D-IQTT (See Figure~\ref{fig:iqtest-sample}). The idea is to use labeled examples to streamline the learned latent representations in order to solve the task.
In this case, we do not assume that we know the camera position for the unlabeled training samples. Ass mentioned earlier, part of the latent vector $\bm{z}$ encodes the actual 3D object (denoted as $\bm{z}_{scene}$) and the remainder estimates the camera-pose (denoted as $\bm{z}_{view}$). For the supervised samples, two additional loss terms were used: (a) a loss that enforces the object component ($\bm{z}_{scene}$) to be the same for both the reference object and the correct answer, (b) a loss that maximizes the distance between the reference object and the distractors. This loss is expressed as:
\begin{equation}
\begin{split}
\begin{aligned}
\mathcal{L}_{\theta} = \frac{1}{2}D_{\theta}(\bm{x}_{ref}, \bm{x}_{ans}) - \frac{1}{2}\sum_{i=1}^{2} D_{\theta}(\bm{x}_{ref}, \bm{x}_{d_i})
\label{eq:app-loss}
\end{aligned}
\end{split}
\end{equation}
where $\bm{x}_{ref}$ is the reference image, $\bm{x}_{ans}$ is the correct answer, $d_i$ denotes the distractors, $D_{\theta}(\bm{x}_1, \bm{x}_2) = (||\bm{z}_{scene}^{\bm{x}_1}- \bm{z}_{scene}^{\bm{x}_2}||_2)^2$ and $\bm{z}^{\bm{x}}=Encoder_{\theta}(\bm{x})$.

During training, we also minimize the mutual information between $\bm{z}_{scene}$ and $\bm{z}_{view}$ to explicitly disentangle both. This is implemented via MINE~\citep{mine}. The strategy of MINE is to parameterize a variational formulation of the mutual information in terms of a neural network:
\begin{equation} \label{Fdiv}
I_{\Theta}(z_{s},z_{v})  =  \sup_{\theta\in\Theta} \EE_{\PJ{z_{s}}{z_{v}}}[\FGen_\theta] - \log\left(\EE_{\PI{z_{s}}{z_{v}}}[e^{\FGen_\theta}]\right).
\end{equation}

This objective is optimized in an adversarial paradigm where $T$, the statistics network, plays the role of the critic and is fed with samples from the joint and marginal distribution. We use this loss to minimize the mutual information estimate in both unsupervised and supervised training iterations. Once the model is trained, we answer 3D-IQTT questions, by inferring the latent 3D representation for each of the four images and we select the answer closest to the reference image as measured by $L_2$ distance on latent representations.
\begin{figure*}[t]
\centering
  \begin{subfigure}[b]{\textwidth}
    \centering
    \includegraphics[width=0.475\columnwidth]{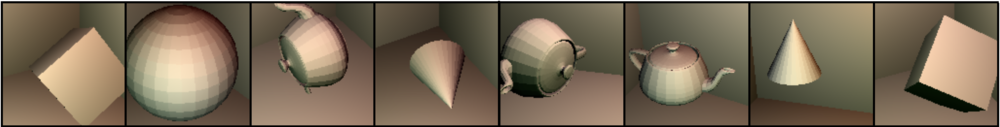}\hspace{0.5cm}
    \includegraphics[width=0.475\columnwidth]{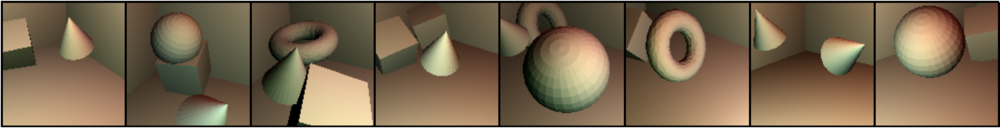}
    \includegraphics[width=0.475\columnwidth]{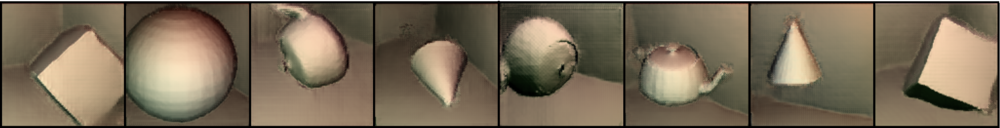}\hspace{0.5cm}
    \includegraphics[width=0.475\columnwidth]{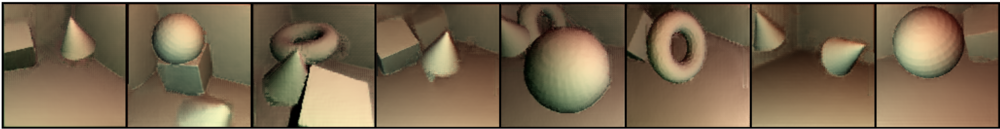}
    \caption{Top: Input images. Bottom: Reconstructed images}
  \end{subfigure}%

  \begin{subfigure}[b]{\textwidth}
    \centering
    \includegraphics[width=0.475\columnwidth]{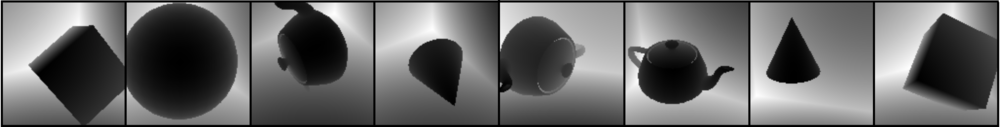}\hspace{0.5cm}
    \includegraphics[width=0.475\columnwidth]{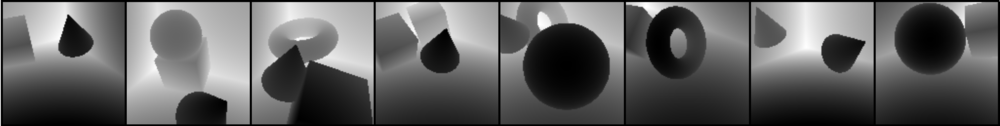}
    \includegraphics[width=0.475\columnwidth]{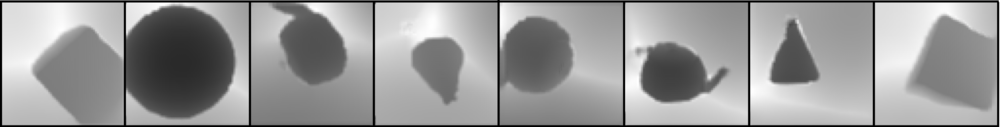}\hspace{0.5cm}
    \includegraphics[width=0.475\columnwidth]{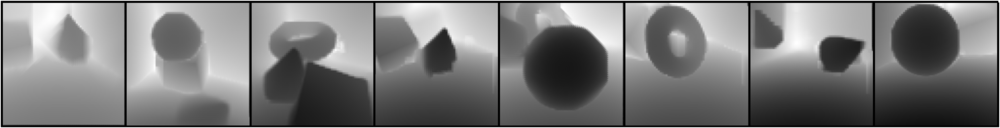}
    \caption{Top: Ground-truth depth maps. Bottom: Reconstructed depth maps}
  \end{subfigure}
  
  \begin{subfigure}[b]{\textwidth}
    \centering
    \includegraphics[width=0.475\columnwidth]{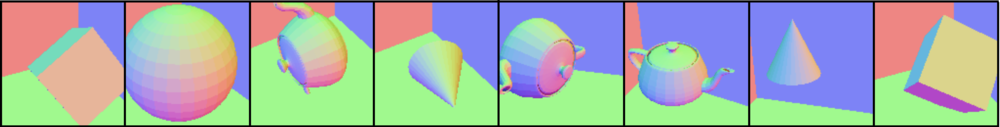}\hspace{0.5cm}
    \includegraphics[width=0.475\columnwidth]{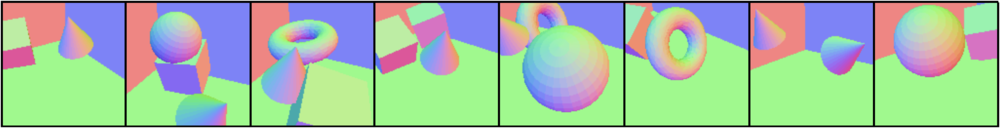}
    \includegraphics[width=0.475\columnwidth]{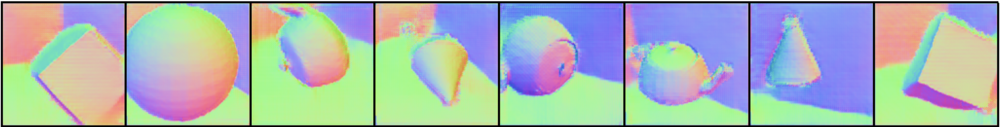}\hspace{0.5cm}
    \includegraphics[width=0.475\columnwidth]{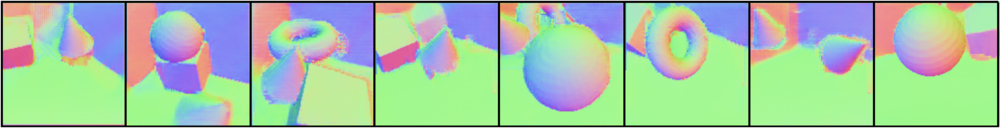}
    \caption{Top: Ground-truth normal maps. Bottom: Reconstructed normal maps}
  \end{subfigure}
  \caption{\textbf{Shape scenes reconstruction.} Pix2Shape reconstruction of single objects in a room (left) and multiple objects into a room (right). On both sides, the ground truths for RGB, depth, and normals are in the upper row, the inferred image, depth and normals are in the respective lower rows. Our model is able to correctly recover the depth and normal of the scenes from a single 2D image.}
  \label{fig:results_reconstruction_shapes}
\end{figure*}

\section{Experimental Setup}
\label{ssec:experimental_setup}

We evaluate Pix2Shape on three different tasks: scene reconstruction, scene generation, and 3D-IQTT.

\subsection{Scene Reconstruction}
The goal of this task is to produce a 2.5D representation (depth and normals) from a given input image. Moreover, we also evaluate if the model can extrapolate to unobserved views of the scene.

For this task we have created two datasets of scene images composed of a room containing one or more objects placed at random positions and orientations. \emph{Shape scenes} dataset is created with rendered images of multiple basic 3D shapes (i.e., box, sphere, cone, torus, teapot etc)placed inside a room. \emph{ShapeNet scenes} dataset is constructed from renderings of multiple objects of different categories from the ShapeNet dataset~\citep{shapenet2015} (i.e., bowls, bottles, mugs, lamps, bags, etc). 

Each 3D scene is rendered into a single $128\times128\times3$ image taken from a camera in a random position sampled uniformly on the positive octant of a sphere containing the room. The probability of seeing the same configuration of a scene from two different views is near zero.

We evaluate the performance of scene reconstruction using three different metrics: (1) Chamfer distance, (2) Hausdorff distance~\citep{hausdorff1949grundz} (on surfels' position), and (3) Mean Squared Error (MSE).

Chamfer distance (CD) gives the average distance from each point in a set to closest point in the other set. For any two point sets $A, B \subset \R ^3$ Chamfer distance is measured using:

\[
\begin{split}
\begin{aligned}
    CD(A, B) =& \dfrac{1}{|A|} \sum_{x \in A}\min_{y \in B}\|x-y\|_2 + \dfrac{1}{|B|} \sum_{x \in B}\min_{y \in A}\|x-y\|_2
\end{aligned}
\end{split}
\]

Hausdorff distance (HD) measures the correspondence of the model's 3D reconstruction with the input for a given camera pose. 
Given two point sets, $A$ and $B$, the Hausdorff distance is, 
$$\max\left\{\max D_H^+(A, B), \max D_H^+(B, A)\right\},$$
where $D_H^+$ is an asymmetric Hausdorff distance between two point sets. E.g., $\max D_H^+(A, B) = \max D(a, B)$, for all $a \in A$, or the largest Euclidean distance $D(\cdot)$, from a set of points in $A$ to $B$, and a similar definition for the reverse case $\max D_H^+(B, A)$. In both the evaluations, we mesaure compare our reconstructed view-centric surfels (3D positions and normals) to the groundtruth surfels.

\begin{table}[b!]
    \centering
    \setlength{\tabcolsep}{2pt}
    \begin{tabular}{lcccc}
      \toprule
                      &   \multicolumn{2}{c}{\textbf{Ours}}  & \multicolumn{2}{c}{\textbf{PTN}}\\\toprule
                      &   \makecell{Shape\\ scenes} & \makecell{ShapeNet\\ scenes}  &   \makecell{Shape\\ scenes} & \makecell{ShapeNet\\ scenes} \\\midrule
      Chamfer distance (CD) &  0.103 & 0.133&0.145&0.181\\
      
      Hausdorff (HD)      &  0.191 & 0.215&0.229&0.254\\
      MSE-depth        & 0.038 & 0.053&0.056&0.103\\
      \bottomrule
    \end{tabular}
    \caption{\label{tab:results_reconstruction} \textbf{Scene reconstruction results.} Evaluation of Pix2Shape on scene reconstruction with Chamfer distance and Hausdorff metric on 2.5D surfels and MSE on the depth maps. Table also compares with view-centric reconstruction of PTN~\cite{Xinchen2016},}
\end{table}

\subsection{Scene Generation}
In the second task we showcase the generative ability of our model by using our generator to sample class conditioned shapes from ShapeNet dataset. We evaluate the 3D scene generation task qualitatively.

\subsection{3D-IQTT}
In the final task we evaluate the 3D understanding capability of the model on 3D-IQTT: a spatial reasoning-based semi-supervised classification task. The goal of the 3D-IQTT is to quantify the ability of our model to perform a 3D spatial reasoning test by using large amounts of the unlabeled training data and a small set of labeled examples.

For this 3D-IQTT task, we generated a dataset where each IQ question consists of a reference image of a Tetris-like shape, as well as three other images, one of which is a randomly rotated version of the reference (see Figure~\ref{fig:iqtest-sample} for an example). The training set consists of 100k questions where only a few are labeled with the information about the correct answer (i.e. either $1\%$ (1k) or $0.2\%$ (200) of the total training data). The validation and test sets each contain 100K labeled questions.  Earlier literature related to 3D-IQTT is elaborated in supplementary material section \ref{app:iqtt}. We evaluate the 3D-IQTT task with the percentage of questions answered correctly. 

More details on experimental setup and evaluation can be found in supplementary material sections \ref{app:camera_sec} and \ref{app:evaluation}.

\begin{figure}[t!]
  \centering
  \begin{subfigure}[b]{\columnwidth}
    \centering
    \includegraphics[width=\columnwidth]{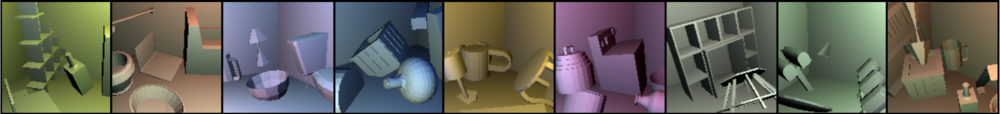}
    \caption{Input images}
  \end{subfigure}
  \begin{subfigure}[b]{\columnwidth}
    \centering
    \includegraphics[width=\columnwidth]{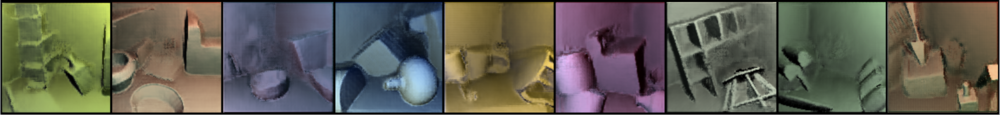}
    \includegraphics[width=\columnwidth]{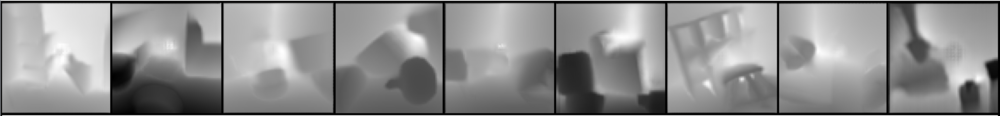}
    \includegraphics[width=\columnwidth]{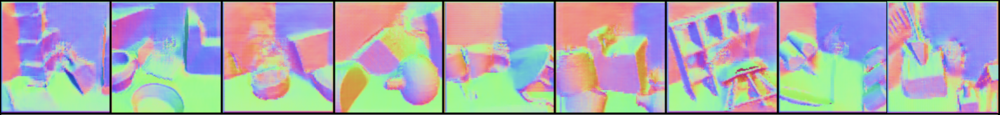}
    \caption{Reconstructed images, depths and normals}
  \end{subfigure}
  \caption{\textbf{ShapeNet scenes reconstruction.} Implicit 3D reconstruction of scenes composed by multiple ShapeNet objects.}
  \label{fig:results_reconstruction_shapenet}
\end{figure}

\begin{figure}[tb]
  \centering
  \begin{subfigure}[b]{\columnwidth}
    \centering
    \includegraphics[width=\columnwidth]{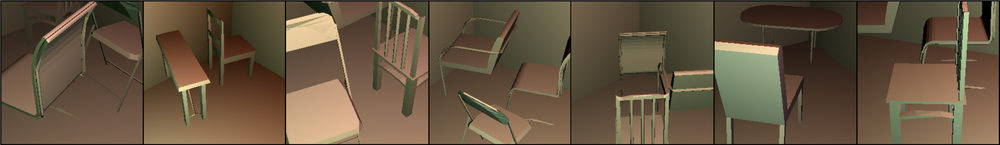}
    \caption{Input images}
  \end{subfigure}
  \begin{subfigure}[b]{\columnwidth}
    \centering
    \includegraphics[width=0.99\columnwidth]{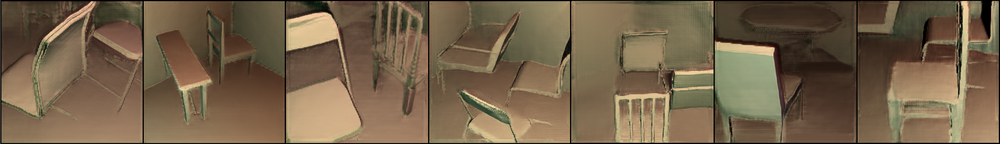}
    \includegraphics[width=0.99\columnwidth]{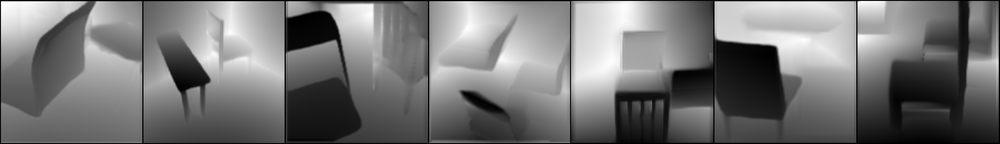}
    \includegraphics[width=0.99\columnwidth]{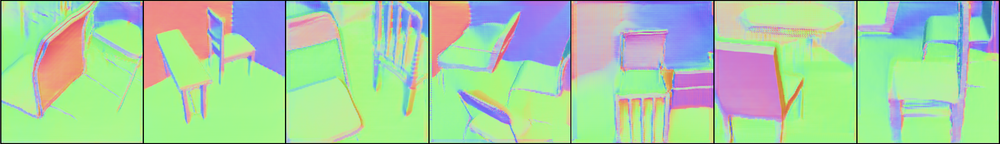}
    \caption{Reconstructed images, depths and normals}
  \end{subfigure}
  \caption{\textbf{ShapeNet $256 \times 256$ scenes reconstruction.} Implicit 3D reconstruction of scenes composed by multiple ShapeNet objects.}
  \label{fig:results_reconstruction_shapenet256}
\end{figure}

\section{Experiments}
\label{sec:experiments}

\subsection{Scene Reconstruction}
\label{ssec:reconstruction}
Figure~\ref{fig:results_reconstruction_shapes} shows the input \textit{shape scenes} data and its corresponding shading reconstructions, along with its recovered depth and normal maps. The depth map is encoded in such a way that the darkest points are closer to the camera. The normal map colors correspond to the cardinal directions (red/green/blue for x/y/z axis respectively). Table~\ref{tab:results_reconstruction} shows a quantitative evaluation of the Chamfer and Hausdorff distances on Shape scene and shapenet scene datasets from a given observed view. The table also depicts mean squared error (MSE) of the generated depth map with respect to the input depth map. The shading reconstructions are almost perfect in this simple dataset. 
Our model successfully learns the depth of the scenes and thereby the relative positions of the surfels. It also estimates the normal maps from the depth consistently. However the absolute distance is not always recovered perfectly.

Figure~\ref{fig:results_reconstruction_shapenet} shows the reconstructions from the model on challenging \textit{ShapeNet scenes} where the number of objects as well as their shape varies. Note how our model is able to handle geometry of varying complexity. Figure~\ref{fig:results_reconstruction_shapenet256} shows reconstructions on $256 \times 256$ resolution scenes(on the right) constructed out of more difficult thin-edged chairs and tables from ShapeNet dataset in random configurations.

\begin{figure}[t]
  \centering
  \includegraphics[width=\columnwidth]{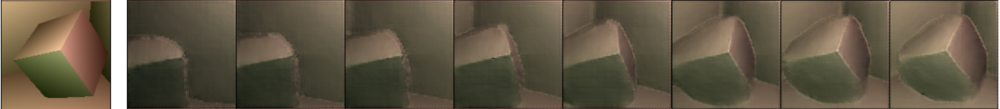}
  \includegraphics[width=\columnwidth]{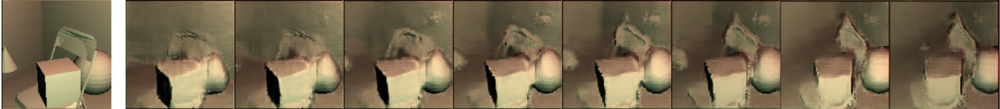}
  \includegraphics[width=\columnwidth]{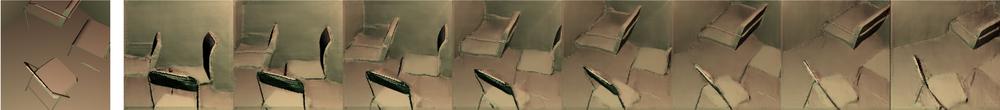}
  \caption{\textbf{Viewpoint reconstruction.} Given a scene (first column), we rotate the camera around it to visualize the unseen parts of the scene. The model correctly infers the unobserved geometry of the objects, demonstrating true 3D understanding of the scene. Videos of these reconstructions can be seen at \href{https://bit.ly/2zADuqG}{https://bit.ly/2zADuqG}.}
  \label{fig:results_reconstruction_view_rotation}
\end{figure}

To showcase that our model can reconstruct unobserved views, we first infer the latent code $\bm{z}_{scene}$ of an image $\bm{x}$ and then decode and render different views while rotating the camera around the scene. Table~\ref{tab:results_reconstruction_view_rotation} shows the Chamfer and Hausdorff distances and MSE loss of reconstructing a scene from different unobserved view angles. As the view angle increases from $0 ^{\circ}$(original) to $80 ^{\circ}$ for \textit{shape scenes} the reconstruction error and MSE tend to increase. However, for the \textit{ShapeNet scenes} the trend is not as clear because of the complexity of the scene and inter-object occlusions. We compare our method with the PTN baseline~\citet{Xinchen2016}. Note that PTN reconstructs the 3D object in voxels explicitly, where as we output a 2.5D representation. Therefore, for a fair comparison we rotate and render per pixel depth map from a desired view and obtain the Chamfer distance with respect to ground truth projection for that view. Figure~\ref{fig:results_reconstruction_view_rotation} qualitatively shows how Pix2Shape correctly infers the scene parts not in view demonstrating true 3D understanding.

\begin{figure}[t]
  \centering
  \begin{subfigure}[t]{0.4\textwidth}
    \centering
  \includegraphics[width=0.88\columnwidth]{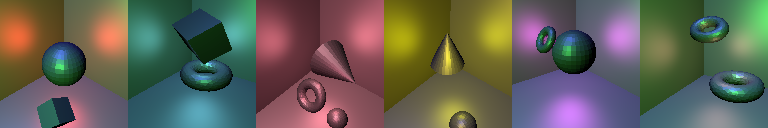}
  \caption{Input images}
  \end{subfigure}
   \begin{subfigure}[t]{0.4\textwidth}
    \centering
  \includegraphics[width=0.88\columnwidth]{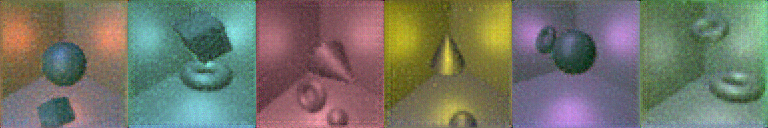}
  \includegraphics[width=0.88\columnwidth]{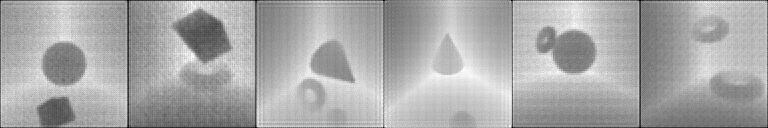}
  \caption{Reconstructed images and depths}
  \end{subfigure}
  \caption{\textbf{Shape scenes reconstruction with specular reflectance.} Pix2Shape reconstruction of multiple objects into a room. The input RGB images are in the upper row, the inferred image and depth are in the respective lower rows.}
  \label{fig:results_reconstruction_specula}
\end{figure}

In all our datasets and further experiments we use diffuse materials with uniform reflectance. The reflectance values are chosen arbitrarily and we use the same material properties for both the input and the generator side. However, our differentiable rendering setup also supports Phong illumination model. As an instance Figure~\ref{fig:results_reconstruction_specula} shows the input \textit{shape scenes} data with specular reflection and its corresponding shading reconstructions, along with its recovered depth.

\subsection{Scene Generation}
\label{sec:experiments_generation}
\begin{figure}[t]
  \centering
  \includegraphics[width=0.48\columnwidth]{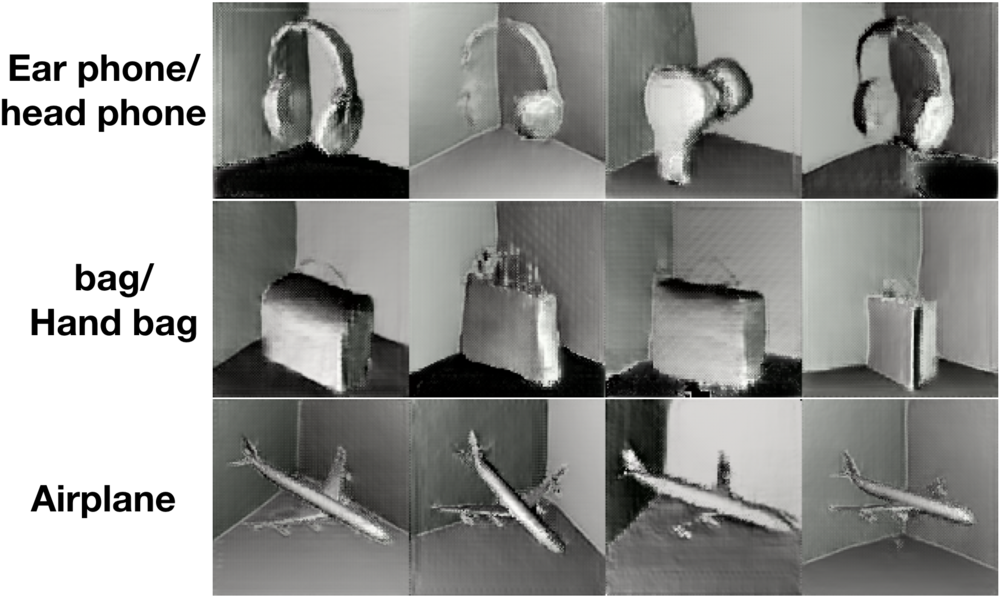}
  \includegraphics[width=0.48\columnwidth]{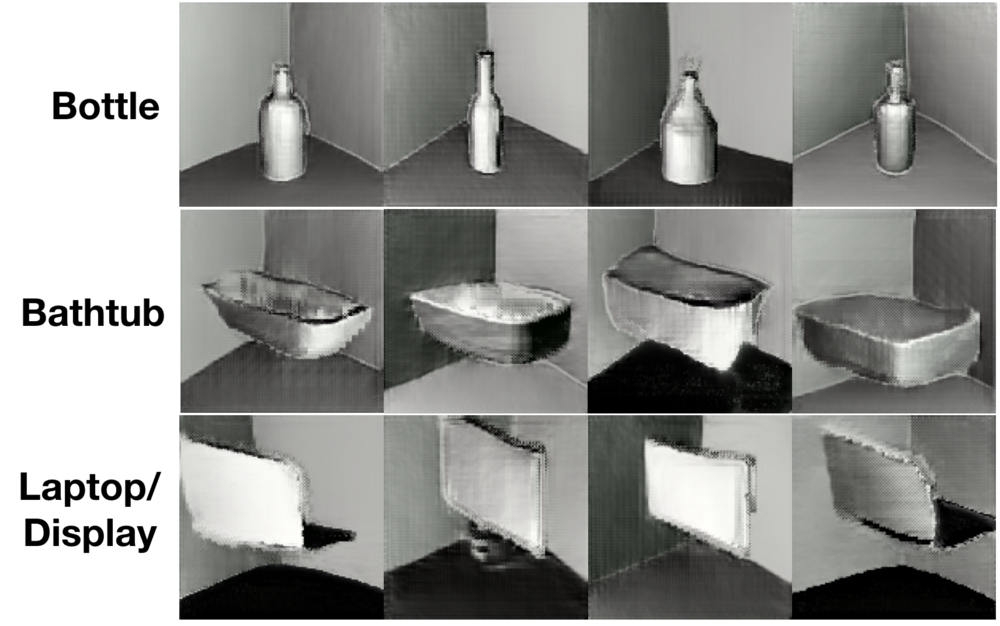}
  \caption{\textbf{Conditional scene generation.} Class-conditionally generated samples for ShapeNet dataset. These images are not part of the training data.}
  \label{fig:results_generation_conditional}
\end{figure}

We trained Pix2Shape on scenes composed of a single ShapeNet object in a room. The model was trained conditionally by giving the class label of the ShapeNet object present in the scene to the decoder and critic networks~\citep{Mirza2014}. Figure~\ref{fig:results_generation_conditional} shows the results of conditioning the decoder on different target classes. Our model was able to generate accurate 3D models for the target class. We can also train the model in an unconditional fashion without giving any object category information (see  supplementary material~\ref{app:unconditional_gen} for more details and results).

\begin{figure}[t]
  \centering
  \includegraphics[width=0.88\columnwidth]{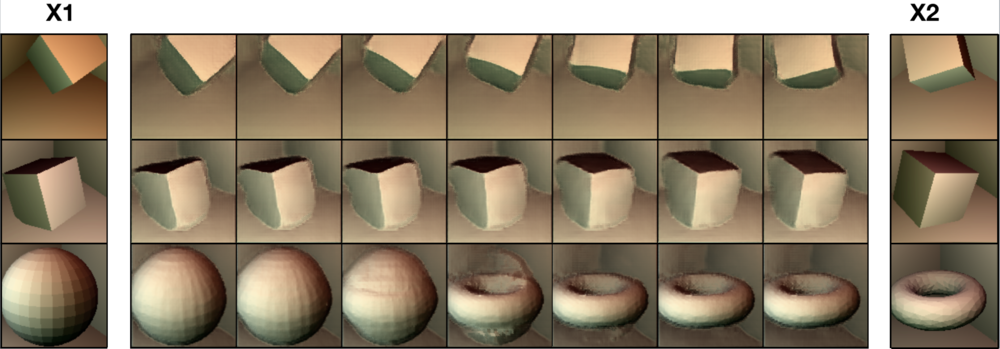}
  \includegraphics[width=0.88\columnwidth]{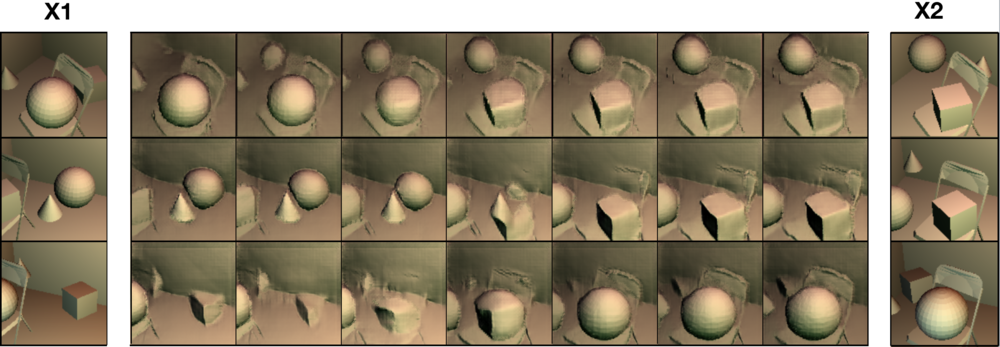}
  \caption{\textbf{Manifold exploration.} Exploration of the learned manifold of 3D representations. Generated interpolations (middle columns) between two images $\bm{x_1}$ and $\bm{x_2}$ (first and last columns).}
  \label{fig:results_generation_manifold_exploration}
\end{figure}

In order to explore the manifold of the learned representations, we selected two images $\bm{x_1}$ and $\bm{x_2}$ from the held out data. We then linearly interpolated between their encodings $\bm{z}_{\bm{1}scene}$ and $\bm{z}_{\bm{2}scene}$ and decoded the intermediary points into their corresponding images using a fixed camera pose. Figure~\ref{fig:results_generation_manifold_exploration} shows this for two different settings. 

\begin{table}[t]
  \setlength{\tabcolsep}{2pt}
  \resizebox{\columnwidth}{!}{%
  \begin{tabular}{c|c|cccc|cccc}
  \hline
  && \multicolumn{4}{c|}{\textbf{Shape scenes}} & \multicolumn{4}{c}{\textbf{Multiple-shape scenes}} \\
  && \includegraphics[height=0.65cm]{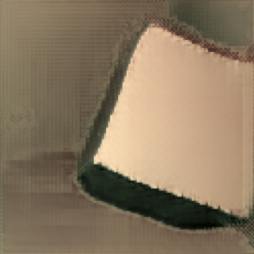} & \includegraphics[height=0.65cm]{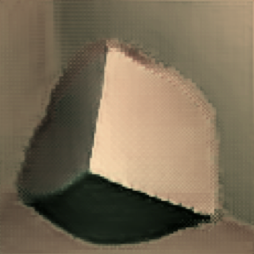}&\includegraphics[height=0.65cm]{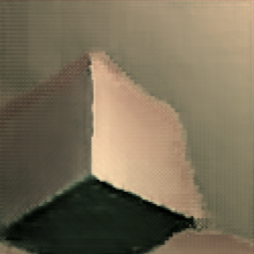} &\includegraphics[height=0.65cm]{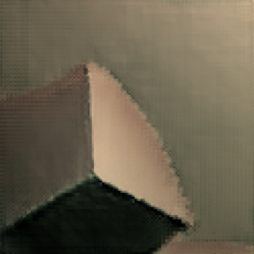} &\includegraphics[height=0.65cm]{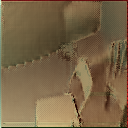} & \includegraphics[height=0.65cm]{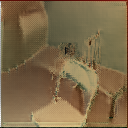}&\includegraphics[height=0.65cm]{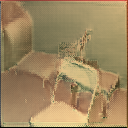} &\includegraphics[height=0.65cm]{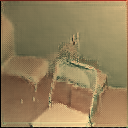}\\
  && 5$^{\circ}$  & 35$^{\circ}$ & 55$^{\circ}$ &80$^{\circ}$ &5$^{\circ}$ & 35$^{\circ}$ & 55$^{\circ}$ & 80$^{\circ}$ \\
  \hline

  &HD & 0.156 & 0.191 & 0.189 & 0.202 & 0.308 & 0.355 & 0.329 & 0.316 \\
  \textbf{Ours}&CD & 0.098 & 0.112 & 0.110 & 0.126 & 0.141 & 0.148 & 0.134 & 0.108 \\
  &MSE   & 0.012 & 0.021 & 0.022 & 0.027 & 0.070 & 0.091 & 0.088 & 0.083 \\
  \hline
  &CD & 0.143 & 0.189 & 0.219 & 0.202 & 0.174 & 0.293 & 0.334 & 0.387 \\
  \textbf{PTN}&MSE   & 0.066 & 0.112 & 0.142 & 0.157 & 0.083 & 0.1969 & 0.1982 & 0.190 \\
  \hline
  \end{tabular}}
  \caption {\textbf{View point reconstruction.} Quantitative evaluation of scene reconstruction for unseen views by extrapolating the view angle from $0 ^{\circ}$(original) to $80 ^{\circ}$. We observe that our method does better when compare to view-centric reconstruction of PTN~\cite{Xinchen2016} Note that PTN model is tuned to perform better for silhouette based single object reconstruction with just plane background. HD is the Hausdorf distance, CD denotes Chamfer Distance and MSE is mean-squared error}
  \label{tab:results_reconstruction_view_rotation}
\end{table}

\begin{table*}[tb!]
    \centering
    \begin{tabular}{@{}llllllll@{}}
        \toprule
        \makecell{Labeled \\ Samples} & \makecell{CNN} & \makecell{Siamese \\ CNN} & \makecell{Human \\ Evaluation} & \makecell{Persp. Transf. Nets \\ \cite{Xinchen2016} } & \makecell{Rezende et al. \\ \cite{Rezende2016} }& \makecell{Pix2Shape \\ \textbf{(Ours)}} \\ \toprule
        0               & 0.3385      &   0.3698     &  0.7329 $\pm$ 0.148  &0.5344 & 0.5202 &  \textbf{0.5519 $\pm$ 0.013}\\
        200             & 0.3350      &   0.3610     & -   &0.6011 & 0.6155 &  \textbf{0.6312 $\pm$ 0.031}\\
        1,000           & 0.3392      &   0.3701     & -   &0.6645 & 0.7001 & \textbf{0.7012 $\pm$ 0.021}\\ \bottomrule
    \end{tabular}
     \captionof{table}{\textbf{3D-IQTT results.} Evaluation on the 3D-IQTT task of our model, two CNN-based baselines, Perspective Transformer Nets and \citep{Rezende2016}. This table also includes comparison with human performance. Although Pix2Shape performs well compared to other baselines, it is still lagging behind the human level by a good margin.}
     \label{tab:results_reconstruction_3d-iqtt}
\end{table*}

\subsection{3D-IQ Test Task}
\label{sec:experiments_understanding}

We trained our model using the aforementioned semi-supervised training described in Section~\ref{3diqtt_method} on the 3D-IQTT task. We compared our model to different baselines listed below and with human performance. 

\paragraph{Human.} We created an online test where 40 random graduate students from our lab answered 20 randomly selected questions from the test set (similar to Figure~\ref{fig:iqtest-sample}). No student had seen these images before. More details can be found in Appendix~\ref{app:3Dhuman}.
\paragraph{CNN.} The first baseline is composed of four ResNet-50 modules~\citep{he2016deep} with shared weights followed by three fully-connected layers and a softmax output for the class label (answer 1 to 3). We trained this CNN only on the labeled samples. The architecture is depicted in the appendix, Figure~\ref{fig:app-baseline-cnn}.
\paragraph{Siamese Network.} Our second baseline is a Siamese CNN with a similar architecture as the previous one but with the fully-connected layers removed. Instead of the supervised loss provided in the form of correct answers, it was trained with contrastive loss~\citep{koch2015siamese}. This loss reduces the feature distance between the reference and correct answer and maximizes the feature distance between the reference and incorrect answers.
\paragraph{Perspective Transformer Nets.} As our third baseline, we used the open source implementation of the Perspective Transformer Nets~\citep{Xinchen2016} to solve the IQTT task using the learnt latent code.
\paragraph{\citet{Rezende2016}:} Since there is no open source code available for this work, we implemented our own interpretation of this work. We were able to reproduce the results from their paper (see appendix \ref{app:rezende} before attempting to use it as baseline for our model.

A more detailed description of the networks and contrastive loss function can be found in the supplementary material~\ref{app:architecture_3diqtt}.

Table~\ref{tab:results_reconstruction_3d-iqtt} shows 3D-IQTT results for our method and baselines. The CNN-based baselines were not able to infer the underlying 3D structure of the data and their results are only slightly better than random guess. The poor performance of the Siamese CNN might be in part because the contrastive loss rewards similarities in pixel space and has no notion of 3D similarity. However, Pix2Shape achieved significantly better accuracy by leveraging the learned 3D knowledge of objects. Our method also outperformed the other 2 baseline approaches, but with a smaller margin.

\begin{table*}[tb!]
    \centering
    \begin{tabular}{ccccccccc}
        \toprule
        \makecell{Adversarial \\ Loss $\mathcal{L}_{ALI}$} & \makecell{Reconstruction \\ Loss $\mathcal{L}_{ALI}$} & \makecell{Mutual Info \\ Loss} & \multicolumn{4}{c}{\makecell{Reconstruction on ShapeNet\\ scenes  (Chamfer dist)}} & \makecell{3D-IQTT \\No Labels } & \makecell{3D-IQTT \\ 1000 labels } \\
        \cmidrule(r){4-7}
        & & & $0^{\circ}$ & $35^{\circ}$ & $55^{\circ}$ &$80^{\circ}$& & \\ \midrule
        \checkmark               &       &   &  1.927 &2.341 & 2.350 & 2.281&0.4024&0.4454 \\
                    & \checkmark      &      & 0.1066 &0.1935 & 0.201&0.1925 &0.3321&0.3789\\
        \checkmark            & \checkmark&  & 0.1120 & 0.1522&0.1483&0.1433 &0.4224&0.4921\\
         & \checkmark   &  \checkmark& - & -&-&-&0.4943&0.6272 \\
        \checkmark & \checkmark   &  \checkmark& - & -&-&-&0.5526&0.7020 \\\bottomrule
    \end{tabular}
     \captionof{table}{\textbf{Loss analysis}. Ablation study of the different losses used to train our model on both reconstruction task and 3D-IQTT task. We evaluate the contribution of each of the objectives in the table. Having the adversarial loss alone negatively affects the reconstructions considerably because, although output images look realistic, they do not match input images very well. Although the reconstruction loss alone does better for reconstruction without view-extrapolation, the performance degrades as we extrapolate to novel views. Note that the reconstruction loss only fixes the indivisibility issues in ALI based models, but considerably affects its generalization ability \citep{li2017alice}}.
     \label{tab:results_Loss_analysis}
\end{table*}
\subsection{Analyzing the Loss Functions}
In this section, we do an ablation study of the different loss functions used to train our model. 
Our final objective is a combination of bi-directional adversarial loss $\mathcal{L}_{ALI}$ and a reconstruction loss $\mathcal{L}_{recon}$. For the 3D-IQTT task we augmented the above losses with a mutual information based objective $I_{\Theta}(z_{s},z_{v})$ to make sure that different parts of the latent code encode distinctive pieces of  information present in a scene. This allows us to disentangle view point and geometry. Table~\ref{tab:results_Loss_analysis} shows our results for both the reconstruction task on the ShapeNet scenes dataset and the 3D-IQTT task when considering, i) only adversarial loss ($\mathcal{L}_{ALI}$); ii) only reconstruction loss($\mathcal{L}_{recon}$); iii) adversarial and reconstruction but not mutual info ($\mathcal{L}_{ALI}$); and ($\mathcal{L}_{recon}$) (note that this does not effect reconstruction task); and, iv) all three ($\mathcal{L}_{ALI}$, $\mathcal{L}_{recon}$ and $I_{\Theta}(z_{s},z_{v})$).

We observe each loss term improves the performance of the model on both the tasks. Using adversarial loss alone is not enough to faithfully reconstruct the surfels. On the other hand we observe that having the reconstruction loss alone affects the performance of the model while extrapolating the shape from unseen views (e.g., view angle $35^{\circ}$ to $80^{\circ}$). However, this scenario yields better performance when reconstructing from the given input view point, i.e., $0^{\circ}$. We also notice that having a reconstruction loss alone affects the quality of the samples generated. We observe that the adversarial loss ($\mathcal{L}_{ALI}$) plays a major role in obtaining  detailed and high quality samples. For the 3D-IQTT task, the role of ($\mathcal{L}_{ALI}$) is more evident. ($\mathcal{L}_{ALI}$) encourages the latent code to learn meaningful representations by constraining the model to match the joint distributions. Results also indicate clearly that skipping mutual information loss degrades the performance of the model on 3D-IQTT task. This is expected because of the mix-up of view information with geometrical information in the latent representation.

\section{Conclusion}
\label{conclusions}
In this paper we propose a generative approach to learn 3D structural properties from single-view images in an unsupervised and implicit fashion. Our model receives an image of a scene with uniform material as input, estimates the depth of the scene points and then reconstructs the input image through a differentiable renderer. We also provide quantitative evidence that support our argument by introducing a novel IQ Test Task in a semi-supervised setup. We hope that this evaluation metric will be used as a standard benchmark to measure the 3D understanding capability of models across different 3D representations. The main drawback of our current model is that it requires the knowledge of lighting and material properties. Future work will focus on tackling the more ambitious setting of learning complex materials and texture along with modelling the lighting properties of the scene.

All code for this project is available at \url{https:// github.com/rajeswar18/pix2shape}. The code we developed in order to reproduce the \citet{Rezende2016} baseline is available at \url{https://github.com/fgolemo/threedee-tools}. 

\bibliographystyle{spbasic}  
\bibliography{biblio}   %

\appendix

\section{Rendering Details}
\label{app:rendering}
\begin{figure}[h]
  \centering
  \begin{subfigure}[b]{0.5\columnwidth}
    \centering\includegraphics[width=\columnwidth]{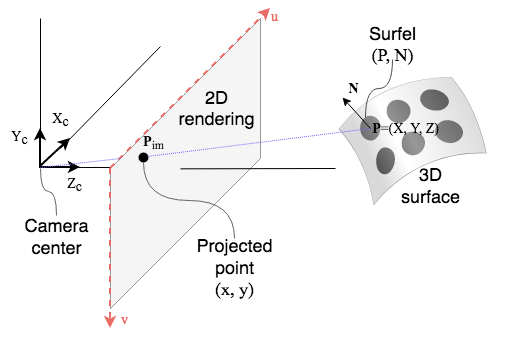}
    \caption{\label{fig:render_raycast}Projection model}
  \end{subfigure}%
  \begin{subfigure}[b]{0.5\columnwidth}
    \centering\includegraphics[width=\columnwidth]{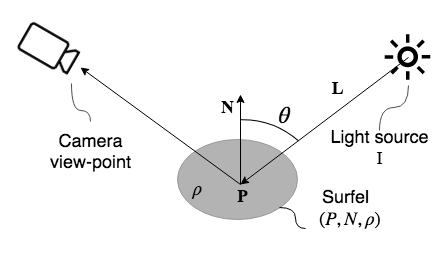}
    \caption{\label{fig:render_shading} Shading model}
  \end{subfigure}%
  \caption{\textbf{Differentiable 3D renderer.} (a) A surfel is defined by its position $P$, normal $N$, and reflectance $\rho$. Each surfel maps to an image pixel $P_{im}$. (b) The surfel's color depends on its reflectance $\rho$ and the angle $\theta$ between each light $I$ and the surfel's normal $N$.}
  \label{fig:render_method}
\end{figure}

The color of a surfel depends on the material reflectance, its position and orientation, as well as the ambient and point light source colors (See Figure~\ref{fig:render_method}b). Given a surface point $P_i$, the color of its corresponding pixel $I_{rc}$ is given by the shading equation:
\begin{equation}
\begin{split}
\begin{aligned}
\label{eq:shading}
  I_{rc} = \rho_i (L_a + \sum_j \frac{1}{k_l \|d_{ij}\| + k_q \|d_{ij}\|^2} L_j \\
  \max\left(0, N_i^T d_{ij} / \|d_{ij}\|\right)),
\end{aligned}
\end{split}
\end{equation}
where $\rho_i$ is the surface reflectance, $L_a$ is the ambient light's color, $L_j$ is the $j^{\mathrm{th}}$ positional light source's color, with $d_{ij} = L_j^{\mathrm{pos}} - P_i$, or the direction vector from the scene point to the point light source, and $k_l$, $k_q$ being the linear and quadratic attenuation terms respectively. Equation~\ref{eq:shading} is an approximation of rendering equation proposed in \citet{Kajiya}.

\section{Architecture}
\label{app:architecture}
Pix2Shape is composed of an encoder network (See Table~\ref{table:architecture_encoder}), a decoder network (See Table~\ref{table:architecture_decoder}), and a critic network (See Table~\ref{table:architecture_critic}). Specifically, the decoder architecture is similar to the generator in DCGAN~\citep{Radford2015} but with LeakyReLU~\citep{mikolov2011empirical} as activation function and batch-normalization~\citep{ioffe2015batch}. Also, we adjusted its depth and width to accommodate the high resolution images accordingly. In order to condition the camera position on the $z$ variable, we use conditional normalization in the alternate layers of the decoder. We train our model for 60K iterations with a batch size of 6 with images of resolution $128 \times 128 \times 3$.

\begin{table}[ht]
\centering
\resizebox{\columnwidth}{!}{%
\begin{tabular}{l|l|c|c|c|l}
\hline
Layer    & Output size             & Kernel  & Str. & BNorm & Activation\\ 
\hlx{vhv}
In $[x, c]$ & $128 \times 128 \times 3$ &              &   &     &      \\
Conv.    & $64  \times 64  \times 85$   & $4 \times 4$ & 2 & Yes & LReLU\\
Conv.    & $32  \times 32  \times 170$  & $4 \times 4$ & 2 & Yes & LReLU\\
Conv.    & $16  \times 16  \times 340$  & $4 \times 4$ & 2 & Yes & LReLU\\
Conv.    & $8   \times 8   \times 680$  & $4 \times 4$ & 2 & Yes & LReLU\\
Conv.    & $4   \times 4   \times 1360$ & $4 \times 4$ & 2 & No  & LReLU\\
Conv.    & $1   \times 1   \times 1$    & $4 \times 4$ & 1 & No  &      \\
\hline
\end{tabular}}
\caption{\textbf{Pix2Shape encoder architecture}}
\label{table:architecture_encoder}
\end{table}

\begin{table}[ht!]
\centering
\resizebox{\columnwidth}{!}{%
\begin{tabular}{l|l|c|c|c|l}
\hline
Layer          & Output size      & Kernel  & Str. & BNorm & Activation\\
\hlx{vhv}
In $[x, c]$ & $131 \times 1$            &              &   &     &     \\
Conv.    & $4   \times 4   \times 1344$& $4 \times 4$ & 1 & Yes & LReLU\\
Conv.    & $8  \times 8   \times 627$  & $4 \times 4$ & 2 & Yes & LReLU\\
Conv.    & $16  \times 16  \times 336$ & $4 \times 4$ & 2 & Yes & LReLU\\
Conv.    & $32  \times 32  \times 168$ & $4 \times 4$ & 2 & Yes & LReLU\\
Conv.    & $64  \times 64  \times 84$  & $4 \times 4$ & 2 & Yes & LReLU\\
Conv.    & $128 \times 128 \times nCh$ & $4 \times 4$ & 2 & Yes &     \\
\hline
\end{tabular}}
\caption{\textbf{Pix2Shape decoder architecture.}}
\label{table:architecture_decoder}
\end{table}

\begin{table}[ht!]
\centering
\resizebox{\columnwidth}{!}{%
\begin{tabular}{l|l|c|c|c|l}
\hline
Layer          & Output size                  & Kernel & Str. & BNorm & Activation\\
\hlx{vhv}
Input $[x, c]$ & $128 \times 128 \times 6$    &          &   &    &    \\
Conv.    & $64  \times 64  \times 85$   & $4 \times 4$ & 2 & No & LReLU\\
Conv.    & $32  \times 32  \times 170$  & $4 \times 4$ & 2 & No & LReLU\\
Conv.    & $16  \times 16  \times 340$  & $4 \times 4$ & 2 & No & LReLU\\
Conv.    & $8   \times 8   \times 680$  & $4 \times 4$ & 2 & No & LReLU\\
Conv. + [z]    & $4 \times 4 \times 1360$ & $4 \times 4$ & 2 & No & LReLU\\
Convolution    & $1 \times 1 \times 1$    & $4 \times 4$ & 1 & No &    \\
\hline
\end{tabular}}
\caption{\textbf{Pix2Shape critic architecture.} Conditional version takes image, latent code $z$ and camera position $c$. }
\label{table:architecture_critic}
\end{table}

\section{Architecture for Semi-supervised experiments}
\label{app:architecture_3diqtt}
Pixel2Surfels architecture remains similar to the previous one but with higher capacity on the decoder and critic. The most important difference is that for those experiments we do not condition the networks with the camera pose to be fair with the baselines. In addition to the three networks we have a statistics network (see Table~\ref{table:architecture_stat}) that estimates and minimizes the mutual information between the two set of dimensions in the latent code using MINE~\citep{mine}. Out of 128 dimensions for $z$ we use first 118 dimensions for represent scene-based information and rest to encode view based info.

\begin{table}[ht]
\centering
\resizebox{\columnwidth}{!}{%
\begin{tabular}{l|l|c|c|c|l}
\hline
Layer          & Output size       & Kernel  & Str. & BNorm & Act.\\
\hlx{vhv}
In $[z[:118], z[118:]]$ & $1 \times 1 \times 128$ &    &   &    & \\
Conv.    & $1  \times 1  \times 256$& $1 \times 1$ & 1 & No & ELU \\
Conv.    & $1  \times 1  \times 512$& $1 \times 1$ & 1 & No & ELU \\
Conv.    & $1  \times 1  \times 1$  & $1 \times 1$ & 2 & No & None\\
\hline
\end{tabular}}
\caption{\textbf{Pix2Shape statistics network architecture.}}
\label{table:architecture_stat}
\end{table}

\begin{figure}[ht]
  \centering
  \includegraphics[width=\columnwidth]{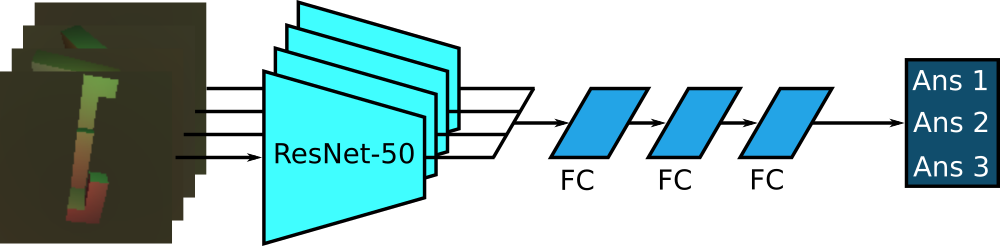}
  \caption{\textbf{3D-IQTT baseline architecture.} The four ResNet-50 share the same weights and were slightly modified to support our image size. ``FC" stands for fully-connected layer and the hidden node sizes are 2048, 512, and 256 respectively. The output of the network is encoded as one-hot vector.}
  \label{fig:app-baseline-cnn}
\end{figure}

The architecture of the baseline networks is shown in Figure~\ref{fig:app-baseline-cnn}. During training we use the contrastive loss~\citep{hadsell2006dimensionality}:
\begin{equation}
\begin{split}
\mathcal{L_\theta}(\bm{x}_{1}, \bm{x}_{2}, y) & = (1-y)\frac{1}{2}(D_\theta(\bm{x}_{1}, \bm{x}_{2}))^2 \\
&  + (y)\frac{1}{2}(max(0,m-D_\theta(\bm{x}_{1}, \bm{x}_{2})))^2\\
D_\theta(\bm{x}_1, \bm{x}_2) & = ||G_\theta(\bm{x}_1)-G_\theta(\bm{x}_2)||_2,
\label{fig:app-baseline-loss}
\end{split}
\end{equation}
where $\bm{x}_1$ and $\bm{x}_2$ are the input images, $y$ is either $0$ (if the inputs are supposed to be the same) or $1$ (if the images are supposed to be different), $G_\theta$ is each ResNet block, parameterized by $\theta$, and $m$ is the margin, which we set to $2.0$. We apply the contrastive loss to the $2048$ features that are generated by each ResNet block.

\section{Material, Lights, and Camera Properties}
\label{app:camera_sec}
\begin{figure}[ht]
  \centering
  \begin{subfigure}[b]{0.45\columnwidth}
    \centering\includegraphics[width=\columnwidth]{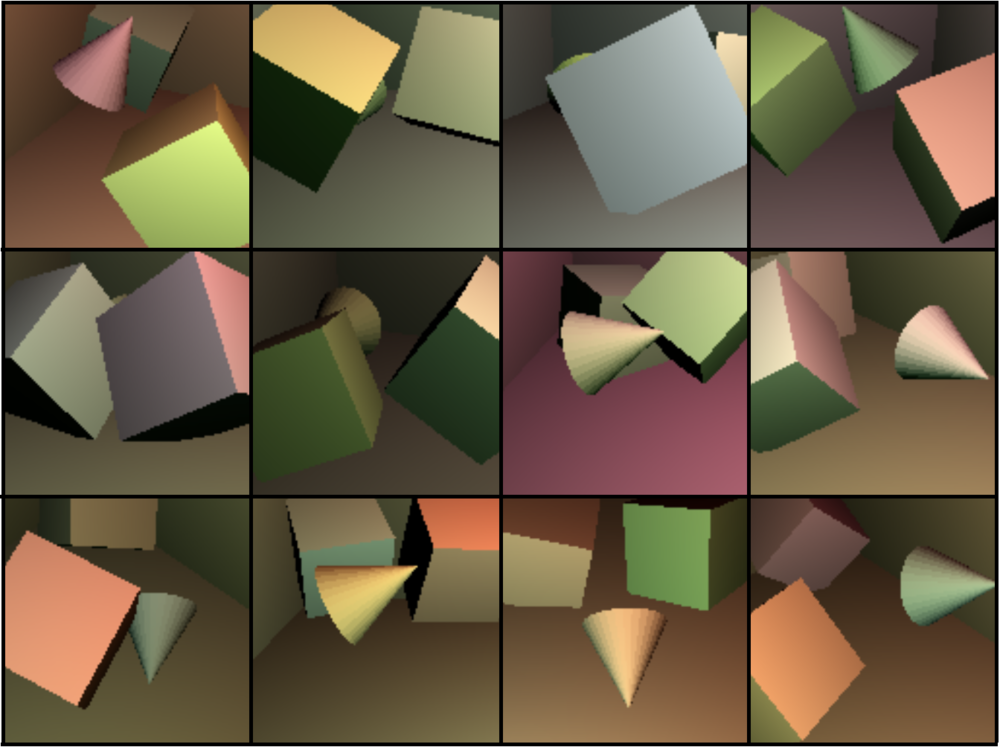}
    \caption{Color input images}
  \end{subfigure}\hfill%
  \begin{subfigure}[b]{0.45\columnwidth}
    \centering\includegraphics[width=\columnwidth]{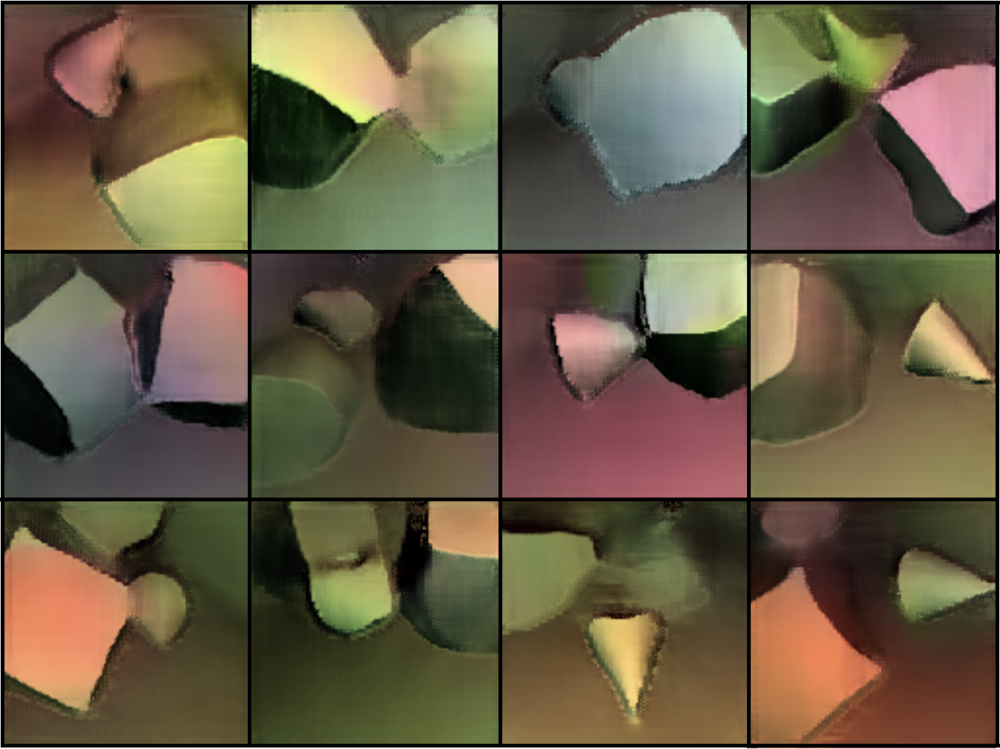}
    \caption{Reconstructed images}
  \end{subfigure}
  
  \begin{subfigure}[b]{0.45\columnwidth}
    \centering\includegraphics[width=\columnwidth]{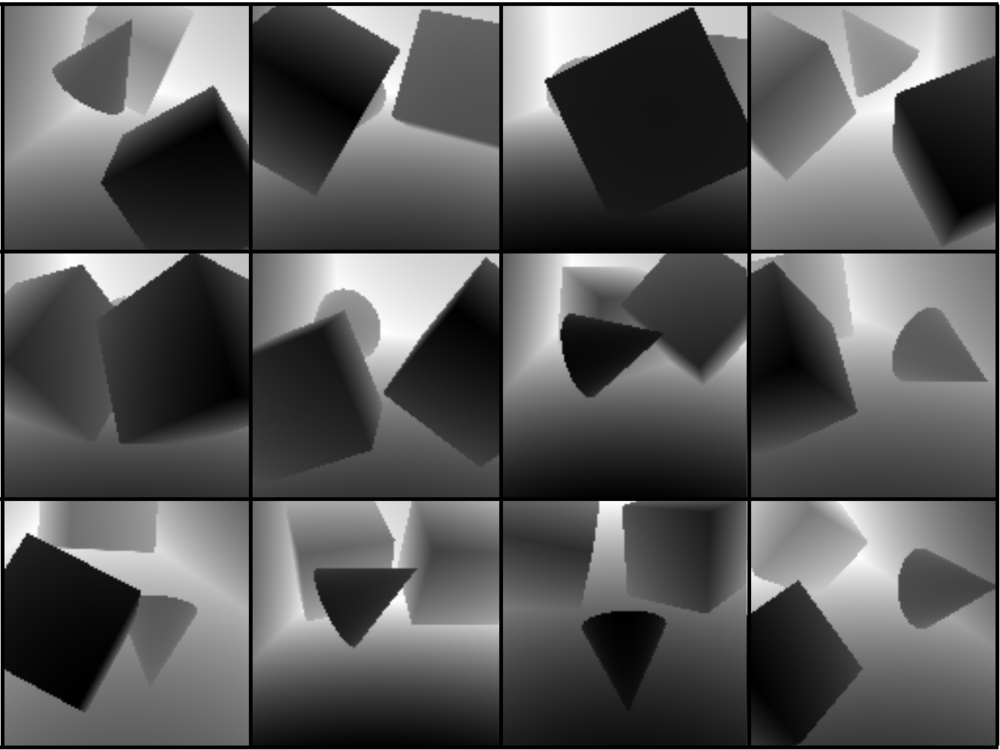}
    \caption{Ground-truth depth}
  \end{subfigure}\hfill%
  \begin{subfigure}[b]{0.45\columnwidth}
    \centering\includegraphics[width=\columnwidth]{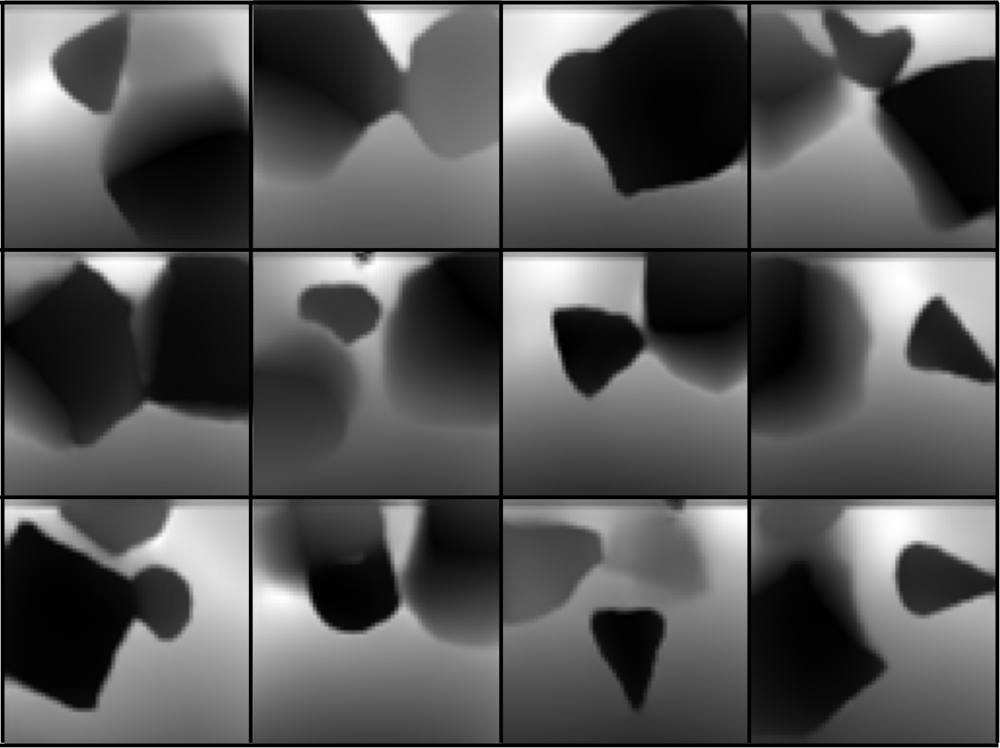}
    \caption{Reconstructed depth}
  \end{subfigure}%
  \caption{\textbf{Learning material along with structure.} The model learns the foreground and background colors separately. }
  \label{fig:results_albedo}
\end{figure}

\paragraph{Material.} In our experiments, we use diffuse materials with uniform reflectance. The reflectance values are chosen arbitrarily and we use the same material properties for both the input and the generator side. Figure~\ref{fig:results_albedo} shows that it is possible to learn reflectance along side learning the 3D structure of the scenes by requiring the model to predict the material coefficients along with the depth of the surfels. The color of the objects depend on both the lighting and the material properties. We do not delve into more details on this, as our objective is to model the structural/geometrical properties of the world with the model. This will be explored further in a later study.

\begin{figure*}[t]
  \centering
  \begin{subfigure}[b]{0.48\textwidth}
    \centering\includegraphics[width=0.99\textwidth,left]{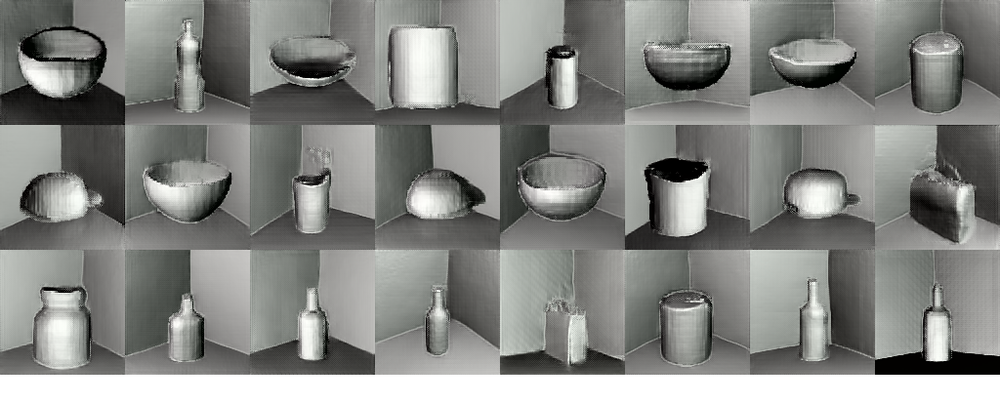}
  \end{subfigure}%
  \begin{subfigure}[b]{0.48\textwidth}
    \centering\includegraphics[width=0.99\textwidth,center]{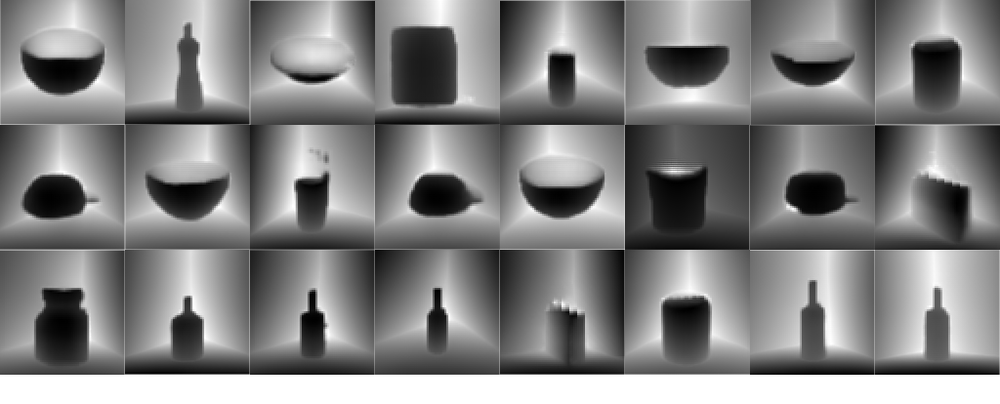}
  \end{subfigure}%
  \caption{\textbf{Unconditional scene generation.} Generated samples from Pix2Shape model trained on ShapeNet scenes. \textbf{Left:} shaded images; \textbf{Right:} depth maps}
  \label{fig:results_generation_unconditional}
\end{figure*}

\paragraph{Camera.} The camera is specified by its position, viewing direction and vector indicating the orientation of the camera. The camera positions were uniform randomly sampled on a sphere for the 3D-IQTT task and on a spherical patch contained in the positive octant, for the rest of the experiments. The viewing direction was updated based on the camera position and the center of mass of the objects, so that the camera was always looking at a fixed point in the scene as its position changed. The focal length ranged between [18 mm and 25mm] in all the experiments and the field-of-view was fixed to 24mm. The camera properties were also shared between the input and the generator side. However, in the 3D-IQTT task we relaxed the assumption that we know the camera pose and instead estimate the view as a part of the learnt latent representation.

\begin{figure}[b]
  \centering
  \begin{subfigure}[b]{0.155\textwidth}
    \centering\includegraphics[width=0.99\textwidth,left]{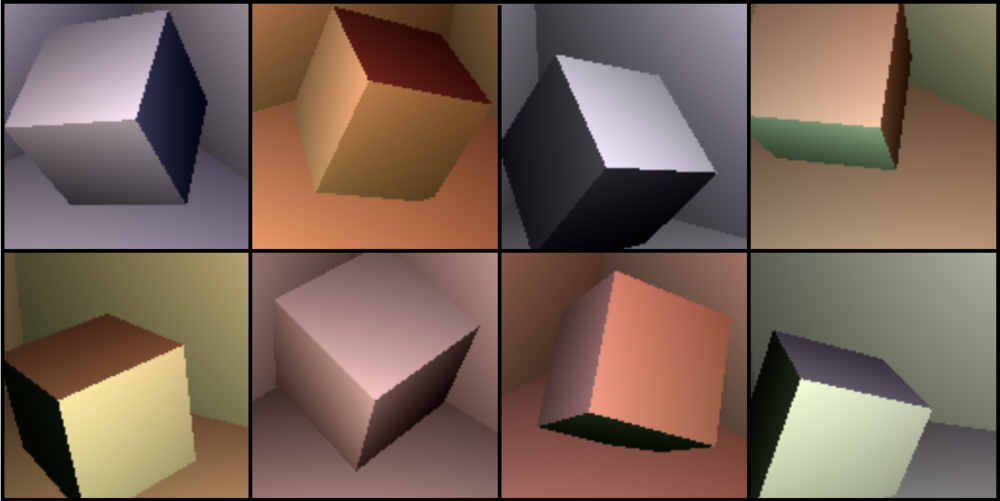}\caption{Input images}
  \end{subfigure}%
  \begin{subfigure}[b]{0.155\textwidth}
    \centering\includegraphics[width=0.99\textwidth,center]{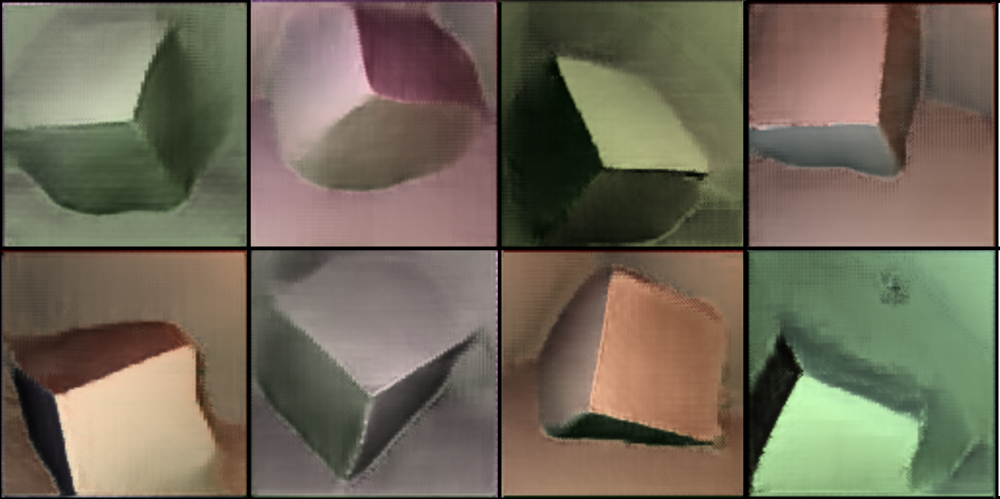}\caption{Reconstructions}
  \end{subfigure}%
  \begin{subfigure}[b]{0.155\textwidth}
    \centering\includegraphics[width=0.99\textwidth,right]{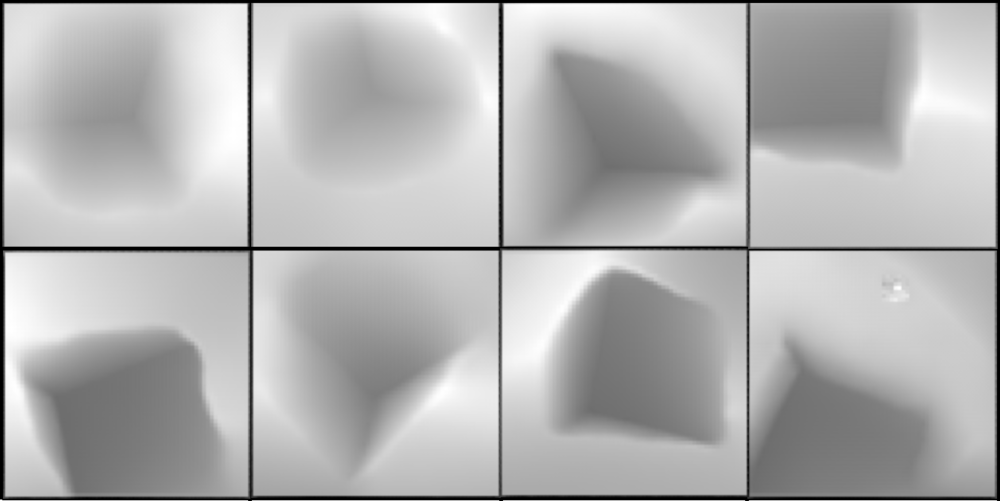}\caption{Recovered depth}
  \end{subfigure}
 
  \caption{\textbf{Random lights configuration.} }
  \label{fig:random_lights}
\end{figure}

\paragraph{Lights.} For the light sources, we experimented with single and multiple point-light sources, with the light colors chosen randomly. The light positions are uniformly sampled on a sphere for the 3D IQTT tasks, and uniformly on a spherical patch covering the positive octant for the other scenes. The same light colors and positions are used both for rendering the input and the generated images. The lights acted as a physical spot lights with the radiant energy attenuating quadratically with distance.
As an ablation study we relaxed this assumption of having perfect knowledge of lights by using random position and random color lights. Those experiments show that the light information is not needed by our model to learn the 3D structure of the data. However, as we use random lights on the generator side, the shading of the reconstructions is in different color than in the input as shown in Figure~\ref{fig:random_lights}.

\section{Unconditional ShapeNet Generation}
\label{app:unconditional_gen}
We trained Pix2Shape on scenes composed of ShapeNet objects from six categories (i.e., bowls, bottles, mugs, cans, caps and bags). Figure~\ref{fig:results_generation_unconditional} shows qualitative results on unconditional generation. Since no class information is provided to the model, the latent variable captures both the object category and its configuration.

\begin{figure*}[t]
  \centering
  \begin{subfigure}[t]{.3\textwidth}
    \centering
    \includegraphics[width=.99\columnwidth]{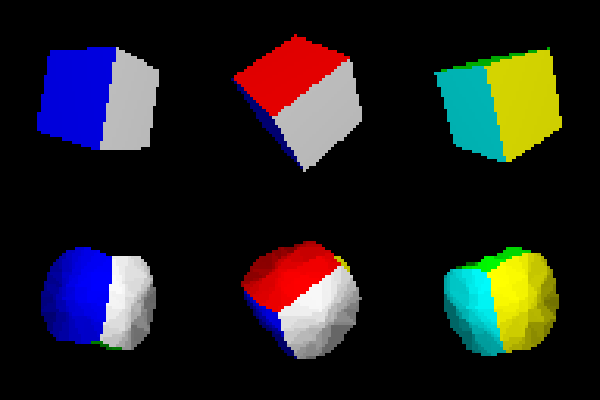}
    \caption{Reproduction of original results.}
    \label{fig:rezende_orig}
  \end{subfigure}\hspace{.5cm}
  \begin{subfigure}[t]{.3\textwidth}
    \includegraphics[width=0.99\columnwidth]{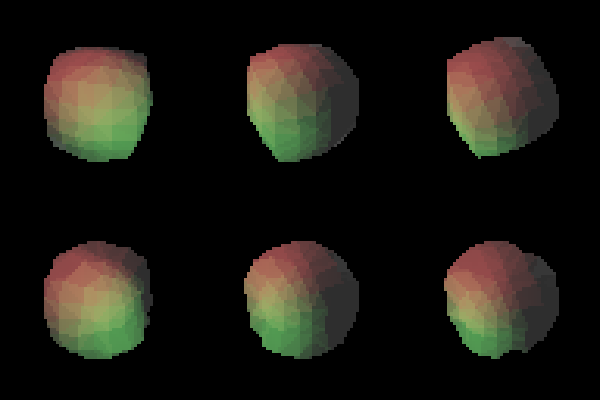}
    \caption{Qualitative results on isolated and centered cube-like shape without background.}
    \label{fig:rezende_cube_nobg}
  \end{subfigure}\hspace{.5cm}
  \begin{subfigure}[t]{.3\textwidth}
    \includegraphics[width=0.99\columnwidth]{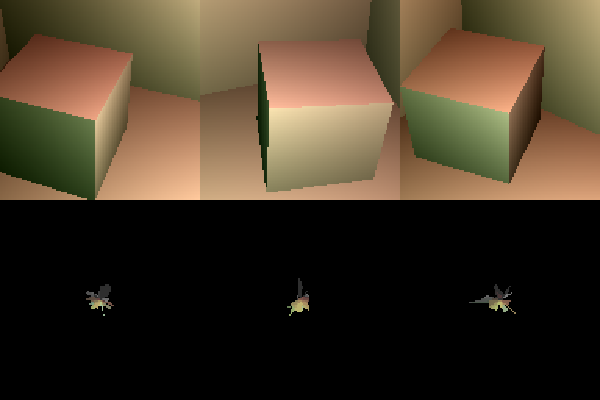}
    \caption{Degenerative results on on full scene.}
    \label{fig:rezende_degen}
  \end{subfigure}
  \caption{\textbf{Reproduction of \citet{Rezende2016} and qualitative results.} Top row: Samples of input images; bottom row: corresponding reconstructed images. We found that with a single centered object, the method was able to correctly reproduce the shape and orientation. However, when the object was not centered, too complex, or there was a background present, the method failed to estimate the shape.}
  \label{fig:rezende}
\end{figure*}
\section{Evaluation of 3D Reconstructions}
\label{app:evaluation}
For evaluating 3D reconstructions, we use the Hausdorff distance~\citep{taha2015efficient} as a measure of similarity between two shapes as in \citet{im2struct}. Given two point sets, $A$ and $B$, the Hausdorff distance is, 
$$\max\left\{\max D_H^+(A, B), \max D_H^+(B, A)\right\},$$
where $D_H^+$ is an asymmetric Hausdorff distance between two point sets. E.g., $\max D_H^+(A, B) = \max D(a, B), \mathrm{for\:all}\,a \in A$, or the largest Euclidean distance $D(\cdot)$, from a set of points in $A$ to $B$, and a similar definition for the reverse case $\max D_H^+(B, A)$.
\section{Ablation study on depth supervision}
As an ablation study, we repeated the experiment that demonstrates the view extrapolation capabilities of our model with depth superrvision. Table \ref{tab:results_reconstruction_view_rotation_depth} depicts the quantitative evaluations on reconstruction if the scenes from unobserved angles.
\begin{table}[b]
  \setlength{\tabcolsep}{2pt}
  \resizebox{\columnwidth}{!}{%
  \begin{tabular}{c|cccc|cccc}
  \hline
  & \multicolumn{4}{c|}{\textbf{Shape scenes}} & \multicolumn{4}{c}{\textbf{Multiple-shape scenes}} \\
  & 5$^{\circ}$  & 35$^{\circ}$ & 55$^{\circ}$ &80$^{\circ}$ &5$^{\circ}$ & 35$^{\circ}$ & 55$^{\circ}$ & 80$^{\circ}$ \\
  \hline
  Hausdorff-F & 0.093 & 0.088 & 0.085 & 0.096 & 0.173 & 0.218 & 0.194 & 0.201 \\
  Hausdorff-R & 0.081 & 0.100 & 0.108 & 0.112 & 0.221 & 0.243 & 0.238 & 0.254 \\
  MSE-depth   & 0.004 & 0.004 & 0.005 & 0.007 & 0.009 & 0.008 & 0.008 & 0.009 \\
  \hline
  \end{tabular}}
  \caption {\textbf{View point reconstruction.} Quantitative evaluation of implicit 3D reconstruction for unseen views by extrapolating the view angle from $0 ^{\circ}$(original) to $80 ^{\circ}$ with depth supervision.}
  \label{tab:results_reconstruction_view_rotation_depth}
\end{table}

\section{3D Intelligence Quotient Task.}
\label{app:iqtt}
In their landmark work, \citet{shepard1971mental} introduced the mental rotation task into the toolkit of cognitive assessment. The authors presented human subjects with reference images and answer images. The subjects had to quickly decide if the answer was either a 3D-rotated version or a mirrored version of the reference. The speed and accuracy with which people can solve this mental rotation task has since become a staple of IQ tests like the Woodcock-Johnson tests~\citep{Woodcock}. We took this as inspiration to design a quantitative evaluation metric (number of questions answered correctly) as opposed to the default qualitative analyses of generative models. We use the same kind of 3D objects but instead of confronting our model with pairs of images and only two possible answers, we include several distractor answers and the subject (human or computer) has to to pick which one out of the three possible answers is the 3D-rotated version of the reference object (See Figure~\ref{fig:iqtest-sample}).
\section{Details on Human Evaluations for 3D IQTT}
\label{app:3Dhuman}
We posted the questionnaire to our lab-wide mailing list, where 41 participants followed the call. The questionnaire had one calibration question where, if answered incorrectly, we pointed out the correct answer. For all successive answers, we did not give any participant the correct answers and each participant had to answer all 20 questions to complete the quiz.

We also asked participants for their age range, gender, education, and for comments. While many commented that the questions were hard, nobody gave us a clear reason to discard their response. All participants were at least high school graduates currently pursuing a Bachelor's degree. The majority of submissions $(78\%)$ were male, whereas the others were female or unspecified. Most of our participants $(73.2\%)$ were between 18 and 29 years old, the others between 30 and 39. The resulting test scores are normally distributed according to the Shapiro-Wilk test $(p<0.05)$ and significantly different from random choice according to 1-sample Student's t test $(p<0.01)$.

\section{Implementation of Rezende et al.}
\label{app:rezende}
With the publication of \citet{Rezende2016}, the authors did not publicly release any code and upon request did not offer any either. When implementing our own version, we attempted to reproduce their results first, which is depicted in Figure~\ref{fig:rezende_orig}. Further, we attempted to use the method for the same qualitative reconstruction of the primitive-in-box dataset as shown in Figure~\ref{fig:results_reconstruction_shapes}. We found that this worked only with one main object and when there was no background (see Figure~\ref{fig:rezende_cube_nobg}). When including the background, applying the same method lead to degenerate solutions (see Figure~\ref{fig:rezende_degen}).

\section {Ablation study of the weighted Mutual-Info loss on 3D-IQTT}
\begin{figure}[t]
   \centering
\begin{tikzpicture}[scale=0.85]
 \begin{axis}[
   legend cell align={left},
   width=1.1\linewidth,
   height=0.8\linewidth,
   xlabel=weight on MI loss ($\lambda$),
   ylabel=3D-IQTT Accuracy,
   grid=major,
   legend pos=south east,
   xlabel near ticks,
   xticklabel style={/pgf/number format/1000 sep=},
   ylabel near ticks,
   xtick=data,
   yticklabel style={
       left,
       /pgf/number format/.cd,
       fixed,
       precision=2,
       /tikz/.cd
   },
   enlarge y limits={value=.1,upper},
   log ticks with fixed point,
   ymin=0.35
 ]
 \addplot[color=green,mark=o] coordinates { (0.1,0.4075) (0.4, 0.4566) (0.6,0.4853) (0.8,0.5130) (1.0,0.5519)  };
 \addplot[color=red,mark=o] coordinates { (0.1,0.4823) (0.4,0.5427) (0.6,0.5854) (0.8,0.6001) (1.0,0.6312)  };

   \legend{No labels, 200 Labels}
 \end{axis}
\end{tikzpicture}
   \caption{\textbf{Study of effect of mutual-information objective on 3D-IQTT performance.} Our model performance is correlated positively to the  the weight on Mutual information term increases}
   \label{fig:my_label}
   \vspace{-5mm}
\end{figure}
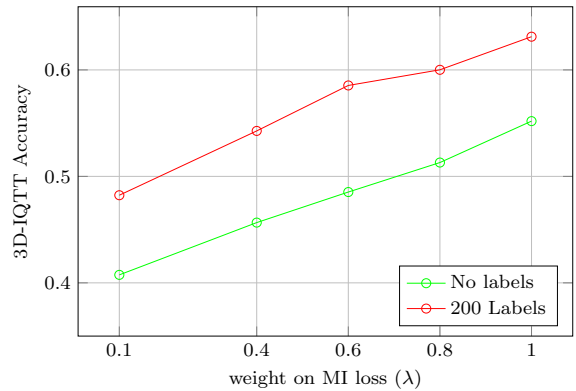

Consider the semi-supervised objective used in algorithm \ref{alg:semi-training}. In this section we do an ablation study on 3D-IQTT performance with the modified form of the equation where Mutual-information loss $I_{\Theta}(z_{scene}, z_{view})$ is weighted by a co-efficient $\lambda$. Plot in Figure~\ref{fig:my_label} indicates the importance of the MI term. Having a good estimate of the view point and disentangling the view information from geometry is indeed crucial to the performance of the IQ task.
\[
L \gets \mathcal{L}_{ALI} + \mathcal{L}_{recon} + I_{\Theta}(z_{scene}, z_{view})
\]

\end{document}